\pdfoutput=1
\documentclass[11pt]{article}

\usepackage{acl}

\usepackage{times}
\usepackage{latexsym}
\usepackage{paralist}

\usepackage[T1]{fontenc}
\usepackage[utf8]{inputenc}

\usepackage{microtype}

\usepackage{inconsolata}
\usepackage{graphicx}
\usepackage[utf8]{inputenc} % allow utf-8 input
\usepackage[T1]{fontenc}    % use 8-bit T1 fonts
\usepackage{hyperref}       % hyperlinks
\usepackage{url}            % simple URL typesetting
\usepackage{booktabs}       % professional-quality tables
\usepackage{amsfonts}       % blackboard math symbols
\usepackage{nicefrac}       % compact symbols for 1/2, etc.
\usepackage{microtype}      % microtypography
\usepackage{xcolor}         % colors

\usepackage{multirow}
\usepackage{adjustbox}
\usepackage{xcolor}
\usepackage{booktabs}

\usepackage{xcolor}
\definecolor{thedarkblue}{RGB}{0,0,120} %104} % 180
\definecolor{mydarkblue}{rgb}{0,0.08,0.45} %ICML dark blue
\definecolor{darkblue}{rgb}{0,0.08,180}
\colorlet{TufteRed}{red!80!black}
\definecolor{theblue}{RGB}{0,0,180}
\colorlet{thered}{TufteRed}
      
\usepackage{microtype}
\usepackage{balance}

\usepackage{booktabs}
\usepackage{tabularx}

\usepackage{amsmath,amssymb,amsthm}

\newcommand{\eat}[1]{\ignorespaces}
\usepackage{comment}

\usepackage{tikz}
\usepackage{verbatim}
\usetikzlibrary{arrows}
\usetikzlibrary{shapes,snakes}
\usetikzlibrary{decorations.pathmorphing} % noisy shapes
\usetikzlibrary{fit}					% fitting shapes to coordinates
\usetikzlibrary{backgrounds}	% drawing the background after the foreground

\usepackage{ragged2e}
\usepackage{multirow}
\usepackage{microtype}
\usepackage{balance}
\usepackage{setspace}

\graphicspath{{./}{./graphics/}}
\newcolumntype{H}{>{\setbox0=\hbox\bgroup}c<{\egroup}@{}}

\newcolumntype{R}[1]{>{\RaggedLeft\arraybackslash}} %p{#1}}
\newcolumntype{L}[1]{>{\RaggedRight\arraybackslash}} %p{#1}}

\newcommand{\eg}{\emph{e.g.}}

\AtBeginEnvironment{pmatrix}{\setlength{\arraycolsep}{2pt}}

\renewcommand{\vec}[1]{\boldsymbol{\mathrm{#1}}}

\DeclareMathOperator{\hugeE}{\mbox{\huge\raise-0.3ex\hbox{E}}}
\DeclareMathOperator{\p}{\mathbb{P}}
\DeclareMathOperator{\hugep}{\mbox{\huge\raise-0.3ex\hbox{$\p$}}}

\providecommand{\vx}{\ensuremath{\vec{x}}}
\providecommand{\vy}{\ensuremath{\vec{y}}}

\usepackage{subfigure}
\usepackage{graphbox}
\usepackage{svg}
\usepackage{nicefrac}
\usepackage{graphicx}

\DeclareMathAlphabet{\mathbcal}{OMS}{cmsy}{b}{n}
\usepackage{amsmath}
\usepackage{comment}

\usepackage{bm}
\usepackage{bbm}
\usepackage{amssymb}
\usepackage{enumitem}

\usepackage{xcolor}
\usepackage{wrapfig,lipsum,booktabs}

\usepackage[strict]{changepage}
\usepackage{float}
\usepackage{framed}
\usepackage[frozencache,cachedir=.]{minted}
\definecolor{formalshade}{rgb}{0.95,0.95,1}

\newenvironment{formal}{%
  \MakeFramed{\advance\hsize-\width\FrameRestore}%
  \noindent\hspace{-4.55pt}% disable indenting first paragraph
  \begin{adjustwidth}{4pt}{7pt}%
}
{%
  \end{adjustwidth}\endMakeFramed%

}

\title{LongLaMP: A Benchmark for Personalized Long-form Text Generation}

\author{Ishita Kumar$^\dagger$, 
Snigdha Viswanathan$^\dagger$,
Sushrita Yerra$^\dagger$,
Alireza Salemi$^\dagger$,\\
\textbf{%
Ryan A. Rossi$^\ddagger$,
Franck Dernoncourt$^\ddagger$, 
Hanieh Deilamsalehy$^\ddagger$,
Xiang Chen$^\ddagger$,}\\
\textbf{%
Ruiyi Zhang$^\ddagger$,
Shubham Agarwal$^\ddagger$,
Nedim Lipka$^\ddagger$,}\\
\textbf{%
Chien Van Nguyen $^{\dagger\dagger}$,
Thien Huu Nguyen $^{\dagger\dagger}$,
Hamed Zamani$^\dagger$}\\
$^\dagger$University of Massachusetts Amherst\\
$^\ddagger$Adobe Research\\
$^{\dagger\dagger}$ University of Oregon\\
\\
The LongLaMP Benchmark: \url{http://LongLaMP-benchmark.github.io}
}

\begin{document}
\maketitle

\begin{abstract}
Long-text generation is seemingly ubiquitous in real-world applications of large language models such as generating an email or writing a review. Despite the fundamental importance and prevalence of long-text generation in many practical applications, existing work on personalized generation has focused on the generation of very short text. To overcome these limitations, we study the problem of \emph{personalized long-text generation}, that is, generating long-text that is personalized for a specific user while being practically useful for the vast majority of real-world applications that naturally require the generation of longer text. In this work, we demonstrate the importance of user-specific personalization for long-text generation tasks and develop the \textbf{Long}-text \textbf{La}nguage \textbf{M}odel \textbf{P}ersonalization (LongLaMP) Benchmark. LongLaMP provides a comprehensive and diverse evaluation framework for personalized long-text generation. Extensive experiments on LongLaMP for zero-shot and fine-tuned language tasks demonstrate the effectiveness of the proposed benchmark and its utility for developing and evaluating techniques for personalized long-text generation across a wide variety of long-text generation tasks. The results highlight the importance of personalization across a wide variety of long-text generation tasks. Finally, we release the benchmark for others to use for this important problem.
\end{abstract}

\begin{figure}[t!]
\centering
% \vspace{-3mm}
\vspace{2mm}
\hspace{-10mm}
\includegraphics[width=1.05\linewidth]{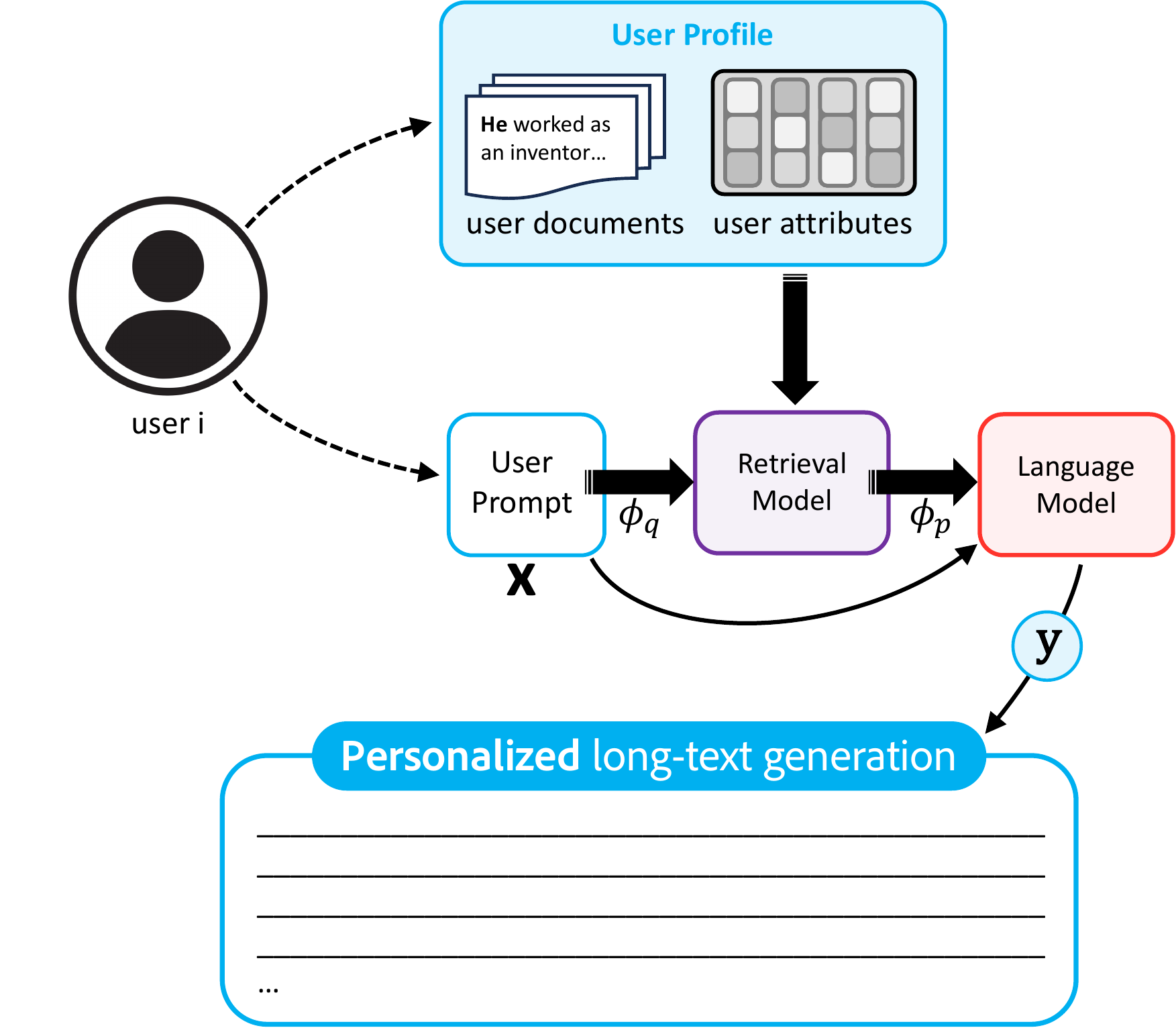}
\caption{%
Overview of the personalized long-text generation framework.
Notably, for generating personalized text for a specific user $i$, the user provides input text $x$, and we leverage their user documents (\eg, review text) and attributes (\eg, ratings) to better personalize the generated text, which is provided as input to the retrieval model.
The output is the personalized long-text generated for that specific user $i$ with the specific input $x$ along with their previous set of user documents and attribute information used to personalize the generated text in terms of style and content.
Note $\phi_q$ and $\phi_p$ are query and prompt construction functions.}
\label{fig:overview}
\vspace{-3mm}
\end{figure}

\section{Introduction}

Personalizing the text generated from Large Language Models (LLMs) has recently attracted significant attention~\cite{Salemi2023LaMPWL,Richardson2023IntegratingSA,richardson2023integrating,Mysore2023PEARLPL,alhafni2024personalized,li-etal-2022-learning-transfer}.
While significant progress has been made for personalization of short-text generation (e.g., generating a subject for an email), the fundamentally more important problem of generating personalized long-text remains relatively unexplored.
Instead of generating a title or email subject, our work focuses on generating the actual content of the paper or text of an email, which is both more complex and useful for a wide variety of applications.
Long-text generation is important for applications like email generation, review generation, and content creation in general, where extended passages of text need to be produced. 
In these contexts, having the ability to generate coherent, contextually relevant long text is crucial~\cite{ji2023survey}.  
Personalizing such long-text generation is particularly challenging due to several factors: maintaining the user's writing style, coherency, and consistency over long outputs; preventing topic drift, and ensuring the generated text stays focused over an extended passage. 

For this new problem of personalized long-text generation, this paper proposes the 
\textbf{Long}-text \textbf{La}nguage \textbf{M}odel \textbf{P}ersonalization (LongLaMP) benchmark,
consisting of 4 diverse personalized tasks: 
(1) Personalized Email Generation,
(2) Personalized Abstract Generation, 
(3) Personalized Review Generation and, 
(4) Personalized Topic Writing to provide a robust and comprehensive evaluation framework for personalized long-text 
generation models. 
For each task, we propose two settings.
(a) User setting that evaluates personalized text generation for new users, we constructed a test set that has no overlap between users from the training and validation set, effectively recreating a cold start scenario. 
(b) Temporal setting that evaluates generating the latest content for previously seen users, we construct test, validation, and training sets with overlapping users in decreasing chronological order. This setting enables the evaluation of the model's ability to adapt and personalize responses for known users, taking into account the user's evolving knowledge and style.

There exist multiple approaches to personalized generation such as fine-tuning an LLM on a per-user basis and latent space representations.
Both approaches suffer from high computational/storage costs, lack of scalability, and potential privacy risks \cite{salemi2024optimization}.
In this work, we investigate a retrieval-augmented generation (RAG) framework for personalized long-text generation.
Our framework, shown in Figure~\ref{fig:overview}, leverages a retrieval model to retrieve relevant user data and integrates it directly into the LLM's input prompts, enhancing the personalization of the generated long-text while maintaining computational efficiency. 
To evaluate the effectiveness of our framework for personalized long-text generation, we performed comprehensive evaluations using our LongLaMP benchmark.
Our findings demonstrate a significant improvement in the ability of these models to generate personalized long-text for individual users.
Specifically, our framework achieves an improvement between 5.7\% to 128\% across various metrics compared to non-personalized baselines.

\smallskip\noindent
This work makes the following contributions:
\begin{itemize}

\item \textbf{Problem Formulation:} We formalize the problem of personalized long-text generation and highlight the importance of personalization for tasks that require the generation of text that is \emph{long} in nature such as an abstract, product review, email, among many others.

\item \textbf{Extensible Open Source Benchmark Environment:} 
We introduce the LongLaMP benchmark consisting of 4 important personalized long-text generation tasks with two settings each.
LongLaMP provides a comprehensive and diverse evaluation framework for personalized long-text generation. It is designed to be easily extended with new models, tasks, and evaluation metrics, among others. We make our benchmark environment publicly available for others to use and extend in their own research \url{ http://LongLaMP-benchmark.github.io}

\item \textbf{Effectiveness: }
We systematically investigate a wide variety of techniques for personalized long-text generation,
and make several interesting findings. The results highlight the importance of personalization for the majority of applications requiring long-text generation.
\end{itemize}

\section{LongLaMP Benchmark}

\subsection{Problem Formulation}
In the context of generative language models, the task of long-text generation can be defined as producing cohesive and contextually relevant textual outputs $y$, which spans multiple sentences, paragraphs, or even pages, conditioned to an input prompt $x$. Personalization for language models is conditioning the textual output $y$ on the input prompt \textit{x} as well as historical and static information about a user \textit{u} (We refer to this user-related information as the \textit{profile}). 

LongLaMP is a benchmark focused exclusively on the challenging task of \textit{personalized long text generation}, where each task is focused on a distinctive domain. For any task $\mathcal{D}$ in LongLaMP, each data entry contains three components: an input prompt $x$, a target output $y$, and a user profile $P_u$. The input prompt $x$, contains information about the personalized task that the user intends to perform. E.g., for writing an email, the input could be the email subject and key points for the body. The target output $y$, is the expected output tailored to the user $u$ (\eg, the generated email personalized for that specific user). The user profile $P_u$ aggregates historical/static data about user $u$. Each entry has task and user-specific attributes. For personalized emails, it could include the subject, dates, etc. of previous emails written.

That said, given a task $\mathcal{D}=\{(x_1, y_1, P_{u_1}), \ldots, (x_n, y_n, P_{u_n})\}$, the main goal of the framework is to maximize the similarity between the generated output $\bar{y}$, and user's expected output $y$ given the input prompt $x$ and the user profile $P_u$.

\subsection{The LongLaMP Benchmark}
The LongLaMP Benchmark consists of four distinct tasks capturing different domains (Table \ref{tab:task-stats}). The tasks vary in audience, purpose, writing style, content type, credibility requirements, length constraints, and structural elements (Table \ref{tab:summary}, Appendix \ref{appendix: data-creation-details}). A rigorous filtering process ensures high quality and utility of the tasks, evaluating challenges and practical applications. Task creation steps are detailed in Appendix \ref{appendix: data-creation-details}.

\paragraph{LongLaMP-1: Personalized Email Completion.}
Email completion can greatly benefit from personalization \cite{trajanovski-etal-2021-text} as email tone varies significantly based on the recipient. This inherent variability in email writing provides a testbed for evaluating the adaptability and personalization of language models. We require models to produce a lengthy email completion, $y$, given an input, $x$, comprising the email subject and partial content. The user profile, $P_u$, consists of previously authored subject-email pairs by that user. To create this task, we utilized the private email collection known as the Avocado Research Email Collection \cite{Oard2015Avocado}. For illustrative purposes, Figure~\ref{benchmark-data-email-completion} in Appendix \ref{appendix:personalized-email-generation-appendix} displays a synthesized example of this long-text email completion task.\footnote{Note that the content shown for Personalized Email Completion in Figure \ref{benchmark-data-email-completion} is synthetically created to preserve confidentiality.}

\paragraph{LongLaMP-2: Personalized Abstract Generation.}
Each researcher has a unique writing style, characterized by factors such as the structure of their arguments, the use of domain-specific language, and the tone they employ making this long-text task challenging. To test the scenarios where the long-text generated output requires domain-specific knowledge for an expert audience we curate the Personalized Abstract Generation as one of the tasks for LongLaMP. The expected output, $y$ is a scientific abstract conditioned on an input, $x$ consisting of the title of the paper and selected keywords from the abstract. The user profile $P_u$ is the previous papers authored by the user. To generate the data samples, we leverage the Citation Network Dataset (V14) \cite{Tang2008ArnetMinerEA}. An example can be seen in Figure \ref{benchmark-data-example-abstraction} in Appendix \ref{appendix:personalized-abstract-generation-appendix}.

\paragraph{LongLaMP-3: Personalized Review Writing.}
Each consumer review reflects the unique perspective and expectations of the reviewer about a product, heavily influenced by personal experiences and specific product features. The style and content of these reviews are adapted to cater to a broad audience of potential buyers. To assess the ability of models to generate long opinionated content, we've established Personalized Review Writing as one of the tasks for LongLaMP. This task is crafted to assess the model's capability to generate a comprehensive, detailed, and long product review, denoted as $y$, from the input $x$ and user profile $P_u$. The input $x$ encompasses the product description, the user's product rating, and an in-depth summary of the user's experience. The user profile $P_u$ consists of other lengthy reviews made by the user described using the review text, a summary of the review, a rating given by the user, and a description of the product. To generate the data samples, we leverage the Amazon Reviews Dataset \cite{Ni2019JustifyingRU}. An example of a data entry, showcasing the long-text nature of the reviews is provided in Figure~\ref{benchmark-data-example-product-review-writinf} in Appendix \ref{appendix:personalized-product-generation-appendix}.

\paragraph{LongLaMP-4: Personalized Topic Writing.} 
Each author on Reddit exhibits a distinctively different writing style based on the subreddit and topic they engage with. Reddit encompasses diverse subreddits dedicated to creative writing, and domain-specific knowledge, many containing linguistic nuances like sarcasm and irony, presenting a comprehensive testbed for evaluating language models' capabilities in personalized long-text generation tasks. This task involves generating the content of a Reddit post, $y$, based on the post's summary, $x$, and the user's previous posts, $P_u$. The user profile, $P_u$ is a compilation of summary-content pairs authored by the user previously. This task is created from the Reddit TL;DR dataset \cite{Vlske2017TLDRMR}. Figure \ref{benchmark-data-topic-generation} in Appendix \ref{appendix:personalized-topic-generation-appendix} displays sample data of the long text generated as reference.

\subsection{Dataset Splits and Evaluation}
\label{sec:2.1}

\paragraph{User Setting:} 
This setting enables evaluation in scenarios where users and items are entirely new to the system.  To facilitate this, we create three distinct sets of users: training, validation, and test datasets. These sets are constructed such that there is no overlap between them.

\paragraph{Temporal Setting:} 
This setting enables the evaluation of change in the linguistic tendencies and style preferences of known users. All users are included in the training dataset, while a subset is incorporated into the validation and test datasets. The user selection is performed in the following manner: all the posts made by a user are sorted chronologically as ${d_1, ..., d_n}$. Based on this ordering, the most recent post is allocated to the test dataset, the penultimate post to the validation dataset, and the third most recent to the training dataset. Documents that precede these selections are aggregated to form the user profile.
Note for the topic writing task, documents are assigned randomly to test, training, and validation sets due to lack of temporal information.

\paragraph{Evaluation:}
\label{sec:2.2}
Following previous works, we use ROUGE-1 and ROUGE-L~\cite{Lin2004ROUGEAP} as task evaluation metrics. 
In addition to that, we also utilize the METEOR score \cite{banerjee2005meteor} as it considers both uni-gram precision and uni-gram recall, accounts for word order differences, and incorporates synonyms and stem mapping into its evaluation resulting in better correlation with human judgment, which can be useful to evaluate the quality of the generated text.
\begin{table*}[h!]
    \centering
    \begin{adjustbox}{max width=\textwidth}    
        \begin{tabular}{lHlcccccccH}
        \toprule
            \textbf{Personalized  Task} & \textbf{Type} & \textbf{Setting} & \textbf{\#train} & \textbf{\#val} & \textbf{\#test} & \textbf{Context Length} & \textbf{Input Prompt Length} & \textbf{Output Length} & \textbf{Profile Size} & \textbf{\#classes}\\
            
            \midrule
            \multirow{2}{*}{Email Completion} & \multirow{2}{*}{text generation} & User & 3286 & 958 & 823 & 191.40 $\pm$ 201.48 & 46.45 $\pm$ 21.45 & 92.59 $\pm$ 60.68 & 85.65 $\pm$ 51.67
            \\
            & & Temporal & 3234 & 833 & 818 & 185.47 $\pm$ 170.41 & 46.75 $\pm$ 21.94 & 92.80 $\pm$ 62.29 & 58.86 $\pm$ 36.23 \\  
            
            \midrule
            \multirow{2}{*}{Abstract Generation} & & User & 13693 & 4560 & 4560 & 144.99 $\pm$  20.69 & 33.82 $\pm$ 5.71 & 144.28 $\pm$ 68.40 & 120.30 $\pm$ 118.81  \\
            & & Temporal & 22822 & 4565 & 4564 & 144.29 $\pm$  20.83 & 34.48 $\pm$ 5.79 & 160.73 $\pm$ 70.74 & 118.17 $\pm$ 118.11 \\
            
            \midrule
            \multirow{2}{*}{Product Review Writing} & \multirow{2}{*}{text generation} & User & 14745 & 1826 & 1822 & 397.37 $\pm$ 144.56 & 119.39 $\pm$ 73.06 & 304.54 $\pm$ 228.61 & 34.39 $\pm$ 57.31  \\
            & & Temporal & 16197 & 1831 & 1784 & 395.69 $\pm$ 142.22 & 121.68 $\pm$ 71.63 &  296.15 $\pm$ 229.13 & 36.75 $\pm$ 60.30  \\
            \midrule
            \multirow{2}{*}{Topic Writing} & \multirow{2}{*}{text generation} & User & 11442 & 2452 & 2453 & 260.86 $\pm$ 124.05 & 28.36 $\pm$ 36.08 & 263.03 $\pm$ 243.34 & 50.39 $\pm$ 2898.60  \\
            & & Random & 16347 & 2452 & 2452 & 261.47 $\pm$ 127.09 & 28.21 $\pm$ 37.76 &  262.58 $\pm$ 241.40 & 43.04 $\pm$ 2542.27  \\
            
        \bottomrule
        \end{tabular}
    \end{adjustbox}
    \caption{%
    Personalized Long-text Generation (LongLaMP) Benchmark
    Statistics.
    Note that context length is the average number of tokens of all the information we use the retrieval search over whereas 
    profile size is the average number of items per user.
    Further, input prompt length denotes the average length of the input given by a user whereas output length refers to the average length of the output written by a user.}
    \label{tab:task-stats}
\end{table*}

\section{Framework}
\label{sec:3}
To personalize the generated outputs, our framework conditions the large language model on the user's profile denoted as $P_u$. Utilizing the entire user profile ($P_u$) may not be feasible due to the high computational costs associated with processing large inputs and the potential for performance degradation. While LLMs can handle larger context windows, studies show performance degrades with longer contexts, as models struggle to robustly utilize extensive input data~\cite{Liu2023LostIT}. 
To overcome these limitations, our personalized long-text generation framework leverages retrieval-augmented generation (RAG), which consists of four components: 
a query generation function $\phi_q$, a retriever $\mathcal{R}$,
a personalized prompt generation function $\phi_p$,  and the large language model itself $\textsc{llm}$. 
The query generation function transforms the input of the user $x_{i}$ into a query ie. $q =\phi_q(x_{i})$. A retrieval model $\mathcal{R}$ that returns the top-$k$ most similar documents from the set $P_u$ for user $u$ based on the query $q$. The $\phi_p$ is the personalized prompt construction function that takes the user text $x_i$ as input and the set of related retrieved documents $\{z_{i1},\ldots,z_{ik}\}$ for user $u$ and outputs a personalized prompt $\bar{x}_{i}$ for the user $u$. Finally the $\textsc{llm}$ that takes in the personalized prompt $\bar{x}_{i}$ and returns the generated output. An overview of the framework is given in Figure \ref{fig:overview}. 

To explore different choices for the retrieval model $\mathcal{R}$, we investigate two approaches: a robust term-matching technique, BM25 \cite{inproceedings}, and a pre-trained dense retrieval model, Contriever \cite{lei-etal-2023-unsupervised}. 
The query generation function $\phi_q$ utilizes the non-templated parts, for example, the title and keywords for the Personalized Abstract Generation task, from the user input $x_{i}$ to create the query $q$. Details for templates of each task are in Figure \ref{fig:overview-prompts}. For the prompt generation function $\phi_p$, we concatenate the task instruction, the input sequence $x_i$, and the retrieved user profile information to construct the final personalized prompt $\bar{x}_i$. Table \ref{tab:prompts-template} details the process to generate $\bar{x}_i$ for each benchmark. 

To fine-tune and evaluate our models, we use a dataset of (prompt, target output) pairs denoted as $(\bar{x_i}, y_i)$, where $\bar{x_i}$ represents the final personalized prompt and 
$y_i$ is the ground-truth text written by the user. The generated text $\bar{y_i}$ from our language model is then evaluated against the actual text $y_i$ using various metrics. 

\section{Experiments}
This section describes our experiments and results demonstrating the utility of the proposed benchmark for personalized long-text generation. Further details on experimental setup is in Appendix \ref{appendix:experimental-setup-extra} and validation results in Appendix \ref{appendix: val-performance}. Additional experiments are in Appendix \ref{appendix:additional-baselines}.

\begin{figure*}[h!]
\centering
\includegraphics[scale=0.35 ]{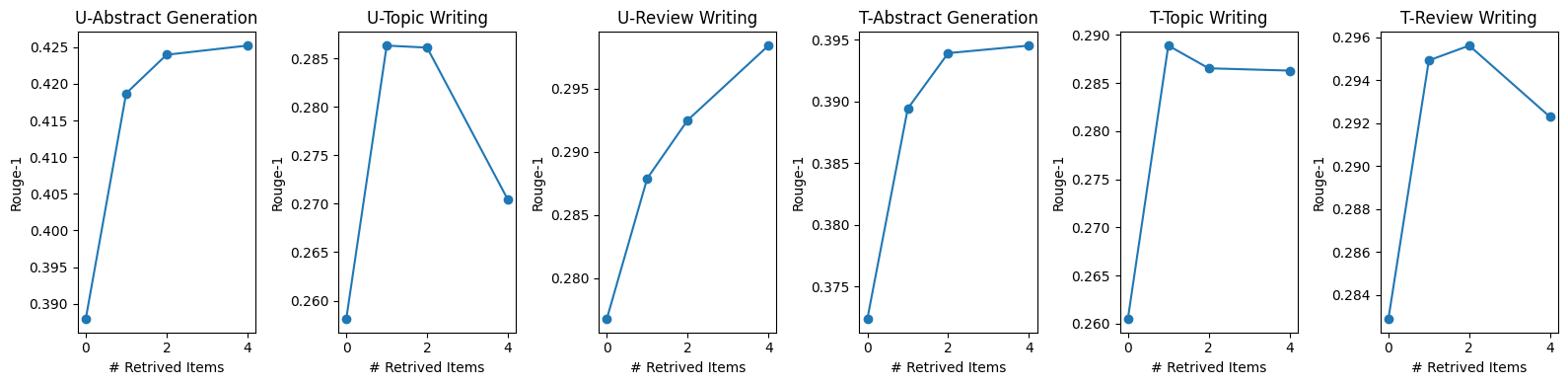}
\caption{The relationship between number $k$ of retrieved profiles.}
% as ROUGE-1 scores}
\label{fig:k_values}
\end{figure*}

\begin{table*}[!t]
    \centering
    \scalebox{0.80}{
        \begin{tabular}{ll cccc cc HH HHHHHH}
        \toprule
        & & \multicolumn{2}{c}{{Non-Personalized}} & 
        \multicolumn{2}{c}{{ Personalized}} &
        \multicolumn{2}{c}{{ Retriever, k}} &
        \\
        \cmidrule(r){3-4}
        \cmidrule(lr){5-6}
        \cmidrule(lr){7-8}
        {{\bf Benchmark Dataset}}  & {{\bf Metric}} & \multicolumn{1}{c}{LLaMA2} & {GPT-3.5\;\;} & {\;\;LLaMA2\;} & {GPT-3.5} & 
        \;\;LLaMA2\;\; &
        \;\;GPT-3.5\;\; &
        \\\midrule
        \multirow{3}{*}{\shortstack[l]{{\bf Personalized}\\{\bf Abstract Generation}}} & ROUGE-1 $\uparrow$ & 0.3508 & 0.3872 & \textbf{0.3917} & \textbf{0.4157} & (Contriever, 4) & (Contriever, 4) & \textbf{0.000} & 0.000 \\
        & ROUGE-L $\uparrow$ & 0.1923 & 0.2207 & \textbf{0.2187} & \textbf{0.2416} & (Contriever, 4) & (Contriever, 4) & \textbf{0.000} & (Contriever, 4) \\
        & METEOR $\uparrow$ & 0.2949 & 0.2530 & \textbf{0.3010} & \textbf{0.2855} & (Contriever, 4) & (Contriever, 4) & \textbf{0.000} & (Contriever, 4) \\
        \midrule
        \multirow{3}{*}{\shortstack[l]{{\bf Personalized}\\{\bf Topic Writing}}} & ROUGE-1 $\uparrow$ & 0.2184 & 0.2773 & \textbf{0.2540} & \textbf{0.2904} & (Contriever, 1) & (Contriever, 1) & \textbf{0.000} & 0.000 \\
        & ROUGE-L $\uparrow$ & 0.1109 & \textbf{0.1397} & \textbf{0.1258} & 0.1368 & (Contriever, 1) & (Contriever, 1) & \textbf{0.000} & 0.000 \\
        & METEOR $\uparrow$ & 0.1635 & 0.1534 & \textbf{0.2004} & \textbf{0.2075} & (Contriever, 1) & (Contriever, 1) & \textbf{0.000} & 0.000 \\
        \midrule 
        \multirow{3}{*}{\shortstack[l]{{\bf Personalized}\\{\bf Review Writing}}} & ROUGE-1 $\uparrow$ & 0.2766 & 0.2746 & \textbf{0.2866} & \textbf{0.2857} & (Contriever, 2) & (Contriever, 4) & \textbf{0.000} & 0.000 \\
        & ROUGE-L $\uparrow$ & 0.1351 & 0.1384 & \textbf{0.1372} & \textbf{0.1450} & (Contriever, 2) & (Contriever, 4) & \textbf{0.000} & 0.000 \\
        & METEOR $\uparrow$ & 0.1620 & 0.1503 & \textbf{0.1865} & \textbf{0.1580} & (Contriever, 2) & (Contriever, 4) & \textbf{0.000} & 0.000 \\
        \midrule
        \multirow{3}{*}{\shortstack[l]{{\bf Personalized}\\{\bf Email Writing}}} 
        & ROUGE-1 $\uparrow$ & 0.1773 & - & \textbf{0.3490} & - & (Contriever, 4) & - & \textbf{0.000} & - \\
        & ROUGE-L $\uparrow$ & 0.1111 & - & \textbf{0.2993} & - & (Contriever, 4) & - & \textbf{0.000} & - \\
        & METEOR $\uparrow$ & 0.1605 & - & \textbf{0.3495} & - & (Contriever, 4) & - & \textbf{0.000} & - \\
        
        \bottomrule
        \end{tabular}
    }
    \caption{The zero-shot personalized results using LLaMA2 and GPT-3.5 models on the test set for the user-based separation setting. 
    The tuned retriever was selected based on the validation performance in Table \ref{table:zero-shot-validation-results-gpt} and Table~\ref{table:zero-shot-validation-results-LLaMA7} in the Appendix.
    Best results are bold. 
    }
    \label{tab:zero-shot-test-results-user-setting}
\end{table*}

\begin{table*}[!t]
    \centering
    \scalebox{0.80}{
       
        \begin{tabular}{ll cccc cc HH HHHHHH}
        \toprule
       
        & & \multicolumn{2}{c}{{Non-Personalized}} & 
        \multicolumn{2}{c}{{ Personalized}} &
        \multicolumn{2}{c}{{ Retriever, k}} &
       
        \\
        
        \cmidrule(r){3-4}
        \cmidrule(lr){5-6}
        \cmidrule(lr){7-8}
       
        {{\bf Benchmark Dataset}}  & {{\bf Metric}} & \multicolumn{1}{c}{LLaMA2} & {GPT-3.5\;\;} & {\;\;LLaMA2\;} & {GPT-3.5} & 
        \;\;LLaMA2\;\; &
        \;\;GPT-3.5\;\; &
        
        \\\midrule
        \multirow{3}{*}{\shortstack[l]{{\bf Personalized}\\{\bf Abstract Generation}}} & ROUGE-1 $\uparrow$ & 0.3515 & 0.3687 & \textbf{0.3697} & \textbf{0.3830} & (Contriever, 4) & (Contriever, 4) & \textbf{0.000} & 0.000 \\
        & ROUGE-L $\uparrow$ & 0.1885 & 0.2094 & \textbf{0.2004} & \textbf{0.2167} & (Contriever, 4) & (Contriever, 4) & \textbf{0.000} & 0.000 \\
        & METEOR $\uparrow$ & \textbf{0.2813} & 0.2314 & 0.2715 & \textbf{0.2477} & (Contriever, 4) & (Contriever, 4) & \textbf{0.000} & 0.000 \\
        \midrule
        \multirow{3}{*}{\shortstack[l]{{\bf Personalized}\\{\bf Topic Writing}}} & ROUGE-1 $\uparrow$ & 0.2193 & 0.2841 & \textbf{0.2487} & \textbf{0.2904} & (Contriever, 1) & (Contriever, 1) & \textbf{0.000} & 0.000 \\
        & ROUGE-L $\uparrow$ & 0.1108 & \textbf{0.1413} & \textbf{0.1226} & 0.1363 & (Contriever, 1) & (Contriever, 1) & \textbf{0.000} & 0.000 \\
        & METEOR $\uparrow$ & 0.1637 & 0.1584 & \textbf{0.1966} & \textbf{0.2077} & (Contriever, 1) & (Contriever, 1) & \textbf{0.000} & 0.000 \\
        \midrule
        
        \multirow{3}{*}{\shortstack[l]{{\bf Personalized}\\{\bf Review Writing}}} & ROUGE-1 $\uparrow$ & 0.2839 & 0.2836 & \textbf{0.2846} &  \textbf{0.2958} & (BM25, 2) & (BM25, 2) & \textbf{0.000} & 0.000 \\
        & ROUGE-L $\uparrow$ & \textbf{0.1395} & 0.1422 & 0.1359 & \textbf{0.1479} & (BM25, 2) & (BM25, 2) & \textbf{0.000} & 0.000 \\
        & METEOR $\uparrow$ & 0.1772 & 0.1612 & \textbf{0.1946} & \textbf{0.1705} & (BM25, 2) & (BM25, 2) & \textbf{0.000} & 0.000 \\
        \midrule
        \multirow{3}{*}{\shortstack[l]{{\bf Personalized}\\{\bf Email Writing}}} 
        & ROUGE-1 $\uparrow$ & 0.1825 & - & \textbf{0.3127} & - & (Contriever, 4) & - & \textbf{0.000} & - \\
        & ROUGE-L $\uparrow$ & 0.1159 & - & \textbf{0.2563} & - & (Contriever, 4) & - & \textbf{0.000} & - \\
        & METEOR $\uparrow$ & 0.1622 & - & \textbf{0.2997} & - & (Contriever, 4) & - & \textbf{0.000} & - \\
        
        \bottomrule
        \end{tabular}
    }
    \caption{The zero-shot personalized results using LLaMA2 and GPT-3.5 models on the test set for the tempora; setting. 
    The tuned retriever was selected based on the validation performance in Table \ref{table:zero-shot-validation-results-gpt} and Table~\ref{table:zero-shot-validation-results-LLaMA7} in the Appendix. The best results are highlighted in bold.}
    \label{tab:zero-shot-test-results-time-setting}
\end{table*}

\begin{table*}[h!]
\centering
\begin{adjustbox}{max width=\textwidth}
\begin{tabular}{lll c ccc ccc}
\toprule
\multirow{2}{*}{\bf Personalized Long-Text} &  \multirow{2}{*}{} & \multirow{2}{*}{} & & \multicolumn{3}{c}{BM25} & \multicolumn{3}{c}{Contriever} \\
\cmidrule(r){5-7} \cmidrule(l){8-10}
{\bf Benchmark Data} & \textbf{Setting} & \textbf{Metric} & Non-Personalized & $k=1$ & $k=2$ & $k=4$ & $k=1$ & $k=2$ & $k=4$ \\
\midrule
\multirow{6}{*}{\shortstack[l]{{\bf Personalized}\\{\bf Abstract Generation}}} &
\multirow{2}{*}{User} & ROUGE-1 $\uparrow$ & 0.3666 & 0.3673 & \textbf{0.3682} & 0.3202 & 0.3673 & 0.3682 & 0.3203 \\
& & ROUGE-L $\uparrow$ & \textbf{0.2442} & 0.2439 & 0.2440 & 0.2099 & 0.2440 & 0.2440 & 0.2100 \\
& & METEOR $\uparrow$ & 0.2251 & 0.2257 & 0.2266 & 0.1993 & 0.2257 & \textbf{0.2266} & 0.1993 \\
\cmidrule{2-10}
& \multirow{2}{*}{Temporal} & ROUGE-1 $\uparrow$ & 0.3471 & \textbf{0.3497} & 0.3440 & 0.2540 & 0.3459 & 0.3441 & 0.2542 \\
&  & ROUGE-L $\uparrow$ & 0.2288 & 0.2317 & 0.2278 & 0.1637 & \textbf{0.2380} &0.2278 & 0.1638 \\
& & METEOR $\uparrow$ & 0.2081 & \textbf{0.2119} & 0.2093 & 0.1562 & 0.2099 & 0.2093 & 0.1562 \\
\midrule
\multirow{6}{*}{\shortstack[l]{{\bf Personalized}\\{\bf Topic Writing}}} & 
\multirow{2}{*}{User} & ROUGE-1 $\uparrow$ & 0.1686 & 0.1712 & 0.1787 & 0.1673 & 0.1722 & \textbf{0.1788} & 0.1672 \\
& & ROUGE-L $\uparrow$ & 0.1195 & 0.1221 & 0.1249 &  0.1132 & 0.1220 & \textbf{0.1250} &  0.1133 \\
& & METEOR $\uparrow$ & 0.0967 & 0.1020 & 0.1038 & 0.0975 & 0.1019 & \textbf{0.1038} & 0.0975 \\
\cmidrule{2-10}
& \multirow{2}{*}{Temporal} & ROUGE-1 $\uparrow$ & 0.1678 & 0.1829 & 0.1853 & 0.1722 & 0.1830 & \textbf{0.1853} & 0.1721 \\
&  & ROUGE-L $\uparrow$ & 0.1201 & 0.1270 & 0.1281 & 0.1155 & 0.1270 & \textbf{0.1282} & 0.1154  \\
& & METEOR $\uparrow$ & 0.0973 & 0.1057 & 0.1072 & 0.1003 & 0.1057 & \textbf{0.1072} & 0.1003 \\
\midrule
\multirow{6}{*}{\shortstack[l]{{\bf Personalized}\\{\bf Review Writing}}} & 
\multirow{2}{*}{User} & ROUGE-1 $\uparrow$ & 0.2039 & 0.2152 & 0.2014 & 0.1918 & \textbf{0.2153} & 0.2013 & 0.1918 \\
&  & ROUGE-L $\uparrow$ & 0.1339 & 0.1383 &  0.1285 & 0.1189 & \textbf{0.1382} & 0.1284 & 0.1188 \\
& & METEOR $\uparrow$ & 0.1030 & 0.1101 & 0.1035 & 0.0995 & \textbf{0.1101} & 0.1035 & 0.0995 \\
\cmidrule{2-10}
& \multirow{2}{*}{Temporal} & ROUGE-1 $\uparrow$ &  0.2070 & \textbf{0.2252} & 0.2139 & 0.2125 & 0.2227 & 0.2138 & 0.2125  \\
&  & ROUGE-L $\uparrow$ & 0.1374 & \textbf{0.1446} & 0.1326 & 0.1208 & 0.1437 & 0.1326 &  0.1206 \\
& & METEOR $\uparrow$ & 0.1092 & \textbf{0.1200} & 0.1143 & 0.1146 & 0.1179 & 0.1143 & 0.1146 \\
\midrule
\multirow{6}{*}{\shortstack[l]{{\bf Personalized}\\{\bf Email Writing}}} &
\multirow{2}{*}{User} & ROUGE-1 $\uparrow$ & 0.1750 & 0.2576 & \textbf{0.3127} & 0.2957 & 0.2628  & 0.3126 & 0.2954 \\
& & ROUGE-L $\uparrow$ & 0.1381 & 0.2226 & 0.2785 & 0.2659 & 0.2261 & \textbf{0.2787} & 0.2657 \\
& & METEOR $\uparrow$ & 0.1387 & 0.2157 & 0.2530 & 0.2456 & 0.2172 & \textbf{0.2530} & 0.2456 \\
\cmidrule{2-10}
& \multirow{2}{*}{Temporal} & ROUGE-1 $\uparrow$ & 0.2356 & \textbf{0.3997} & 0.3962 & 0.3837 & 0.3996 & 0.3958 & 0.3831 \\
&  & ROUGE-L $\uparrow$ & 0.1944 &  0.3615 & 0.3584 & 0.3425 & \textbf{0.3619} & 0.3582 & 0.3424 \\
& & METEOR $\uparrow$ & 0.1930 & 0.3416 & 0.3387 &  0.3255 & \textbf{0.3416} & 0.3387 & 0.3255 \\
\bottomrule
\end{tabular}
\end{adjustbox}
\caption{Personalized long-text generation results on a fine-tuned language model: FlanT5-base on the test set. Note k denotes the retrieved item count for a specific user for personalizing the generation.}
\label{table:finetuning-test-results}
\end{table*}

\subsection{Experimental Setup}
For zero-shot experiments, we leverage GPT-3.5\footnote{GPT-3.5-turbo-viet, accessed through the Azure OpenAI API (version 2023-07-01-preview).} (size unknown) \cite{achiam2023gpt} and LLaMA-2\footnote{Llama-2-7b-chat-hf, accessed through the vLLM library: \url{https://github.com/vllm-project/vllm}} (7B parameters) \cite{Touvron2023LLaMAOA} as LLMs. 
We employ nucleus sampling with temperature 0.8 \cite{holtzman2019curious} as the decoding technique. 
For fine-tuning experiments, we employed FLAN-T5 base \cite{longpre2023flan}. The model was implemented using the Huggingface \texttt{transformers} library. We use beam search \cite{freitag-al-onaizan-2017-beam} with size of 4 for decoding. All experiments were evaluated using the \texttt{evaluate} library. The experiments were performed on Nvidia RTX8000 and A100 GPUs with 49GB and 80GB of GPU memory and 128GB of CPU memory for a max of 3 days on each experiment. 
All the results are reported on one run.

\subsection{Zero-shot Results}
For zero-shot experiments, GPT-3.5 and LLaMA2 are utilized, and the evaluation metrics used are ROUGE-1, ROUGE-L, and METEOR. To evaluate the results, we assessed the generated output corresponding to each input against the expected output, as described in Section \ref{sec:3}.
For our experiments, we employed two widely-used retrievers: BM25 \cite{inproceedings}, a classical bag-of-words retriever, and Contriever \cite{lei-etal-2023-unsupervised} a more recent neural retriever. We further studied the effect of varying the number of retrieved profiles (\textit{k}) on the generated output. Detailed results of these experiments, conducted on both user and temporal settings for the validation sets, are provided in Appendix \ref{appendix: val-performance}.

\subsubsection{Personalized Email Completion}
\textbf{User Setting:}
Personalized results show significant improvement over non-personalized ones using LLaMA2\footnote{Given that this task is based on a private dataset, the experiments were not conducted using GPT-3.5.}  (Table \ref{tab:zero-shot-test-results-user-setting}). The configuration with the best results is Contriever with 4 profiles retrieved. Performance across all metrics improves notably, with ROUGE-L showing a substantial gain of 169.40\% using the LLaMA2 model, as illustrated in the Table~\ref{tab:gain-results-user-setting} in Appendix \ref{appendix: val-performance}.
\\\textbf{Temporal Setting:} The outcomes achieved through utilizing LLaMA2 exhibit improvement in performance as depicted in Table \ref{tab:zero-shot-test-results-user-setting}. Contriever emerges as the top retriever with 4 retrieved profile entries. All metrics show a remarkable improvement in overall gain of above 50\% as displayed in Table \ref{tab:gain-results-temporal-setting} in Appendix \ref{appendix: val-performance}.

\subsubsection{Personalized Abstract Generation}
\textbf{User Setting:}
Personalized results outperform the non-personalized results using both the models as shown in Table \ref{tab:zero-shot-test-results-user-setting}. ROUGE-L scores exhibit the highest overall gain of 12.09\% across models as depicted in Table \ref{tab:gain-results-user-setting} in Appendix \ref{appendix: val-performance}. Contriever, with 4 retrieved profiles performs best.
\\\textbf{Temporal Setting:}
Improvements across all metrics, with a slight degradation in METEOR score using LLaMA2 are observed, as displayed in Table \ref{tab:zero-shot-test-results-time-setting}.
ROUGE-1 and ROUGE-L scores demonstrate substantial overall gains of 4.52\% and 4.91\% respectively the models as seen in Table \ref{tab:gain-results-temporal-setting} in Appendix \ref{appendix: val-performance}. Contriever, with 4 retrieved profiles performs best.

\subsubsection{Personalized Review Writing}
\textbf{User Setting:}
Personalized results surpass the non-personalized results across all metrics while using both LLaMA2 and GPT-3.5 models as depicted in Table \ref{tab:zero-shot-test-results-user-setting}. Contriever is the best-performing retriever for both models with the number of retrieved profile entries as 2 and 4 respectively. METEOR score shows a substantial improvement with an overall gain of 10.12\% across both models as seen in Table \ref{tab:gain-results-user-setting} in Appendix \ref{appendix: val-performance}.

\textbf{Temporal Setting:}
Personalized results outperform the non-personalized results for GPT-3.5 and LLaMA2 models with the exception of a slight decline in ROUGE-L score while using the LLaMA2 as displayed in Table \ref{tab:zero-shot-test-results-time-setting}. BM25 is the best-performing retriever for both models with 2 retrieved profile entries. The METEOR score demonstrates a significant improvement, with an overall gain of 7.79\% across both the models as displayed in Table \ref{tab:gain-results-temporal-setting} in Appendix \ref{appendix: val-performance}.

\subsubsection{Personalized Topic Writing}
\textbf{User Setting:}
The results demonstrate substantial improvement in performance across all metrics while using the LLaMA2 model as displayed in Table \ref{tab:zero-shot-test-results-user-setting}. Except for a very small decline in the ROUGE-L score, performance improvements can be seen across all other metrics
while using the GPT-3.5 model. 
Contriever is seen to be the best-performing retriever for both models with 1 retrieved profile entry. 
METEOR score exhibits a notable overall performance gain of 28.92\% across all models as seen in Table \ref{tab:gain-results-user-setting} in Appendix \ref{appendix: val-performance}. 

\textbf{Temporal Setting:}
The results indicate significant enhancements in performance across all metrics when utilizing the LLaMA2 model as displayed in Table \ref{tab:zero-shot-test-results-time-setting}. GPT-3.5 exhibits a slight deterioration in the ROUGE-L score, however, shows performance improvements for ROUGE-1 and METEOR scores. Contriever is shown to be the best retriever for both the models with 1 retrieved profile. METEOR score demonstrates the highest overall performance gain of 25.61\% across all models as seen in Table \ref{tab:gain-results-temporal-setting} in Appendix \ref{appendix: val-performance}.

\subsection{Fine-tuning Results}
The experimental setup for fine-tuning follows the zero-shot setup described earlier. The results shown in Table \ref{table:finetuning-test-results}, indicate that personalization with fine-tuning yields improvement over non-personalized results. Overall gains for both the settings are discussed in Table \ref{table:gain-results-fine-tuning} in Appendix \ref{appendix: val-performance}.

\textbf{Personalized Email Completion:} For this task, both settings show improvements, with ROUGE-L showing the highest gains. In the user setting, the best performance is achieved by retrieving 2 profiles, resulting in a 101.8\% increase in ROUGE-L scores, while the temporal setting sees an 86.2\% gain. Contriever generally excels in ROUGE-L, and BM25 performs better in ROUGE-1, with both models showing similar METEOR scores.

\textbf{Personalized Abstract Generation:} The user setting shows better results in ROUGE-1 and METEOR, but a slight decrease in ROUGE-L. Optimal performance is achieved by retrieving two profiles, where both retrievers performs equally well in ROUGE-1 and METEOR metrics. In the temporal setting, personalized outcomes improve across all metrics, with ROUGE-L increasing by 4.02\%. Retrieving one profile yields the best results, with BM25 excelling in ROUGE-1 and METEOR, while Contriever leads in ROUGE-L.

\textbf{Personalized Review Writing:} Personalized results outperform non-personalized results across all metrics, with the best results obtained by retrieving one profile. In the user setting, METEOR shows the highest gain of 6.89\%, with Contriever performing slightly better in ROUGE-1 and ROUGE-L, and equally in METEOR. In the temporal setting, METEOR achieves the highest gain of 9.89\%, with both models performing equally in METEOR, Contriever showing slightly better results in ROUGE-L, and BM25 outperforming in ROUGE-1.

\textbf{Personalized Topic Writing:} For this task, both settings show improved personalization results with two profiles retrieved. In the user setting Contriever performs slightly better in ROUGE-1 and ROUGE-L, and BM25 performs better in METEOR. METEOR sees the highest gain of 7.34\%. In the temporal setting, the highest gains are observed in ROUGE-1 at 10.43\%. Contriever shows slight advantages in ROUGE-1 and ROUGE-L, while both models perform similarly in METEOR.

\subsection{Hyperparameter Sensitivity }
We analyzed the impact of varying the number of profiles provided to the personalized LLM (Figure \ref{fig:k_values}). The proposed framework improved performance across all benchmarks compared to non-personalized baselines. For LongLaMP-2, more profiles generally improved abstract quality but with diminishing returns. For LongLaMP-4, ROUGE scores increased from k=0 to k=1 but declined after, suggesting too many profiles degraded performance. In LongLaMP-3, the user setting benefited from more profiles, but the temporal setting plateaued or declined after a certain count. Carefully tuning the profile count based on the task, setting, and model architecture is crucial, as a one-size-fits-all approach may be ineffective.

We also compare our method to two baselines: Non-Personalized Random (random retrieval from all users' profiles) and Personalized Untuned (random retrieval from target user's profiles). Results are in Tables \ref{tab:additional-baselines-user-setting-llama} and \ref{tab:additional-baselines-temporal-setting} in Appendix \ref{appendix:additional-baselines}. For the temporal setting of our benchmark, we experiment with an additional retriever, called Recency detailed in Appendix \ref{appendix:additional-baselines}. Results are in Table \ref{tab:recency-test-setting}.

\section{Related Work}
Domain-specific personalization aims to tailor models to individual users or specific domains. It has been explored across various areas such as product review generation \cite{li2019towards,li2020knowledge}, dialog agents \cite{zhang2018personalizing,mazare-etal-2018-training,10.1145/1401890.1402008}, sentiment analysis \cite{el2023sentiment,mireshghallah-etal-2022-useridentifier}. \citet{Salemi2023LaMPWL} introduce a benchmark for evaluating personalized LLMs using the RAG approach. RAG architectures are increasingly adopted for personalized agents  \cite{wang2024unims,quidwai2024rag}, due to their ability to retrieve relevant passages to augment prompts.

Along with retrieval, various approaches have been used to personalize LLMs, including summarizing user profile items~\cite{Richardson2023IntegratingSA}, models trained or prompted for capturing user style~\cite{Mysore2023PEARLPL,alhafni-etal-2024-personalized}, automatic prompt generation tailored to individual users\citet{Li_2024}, and training retrieval models using reinforcement learning to personalize LLMs \cite{salemi2024optimization} to name a few. While these existing techniques have made progress, they have been focused on short text generation. 
Our work studies the personalized long-text generation problem that is of more practical importance with a wide variety of applications. \citet{li2023teach} explores the scenario of personalized long text generation, by proposing a multistage framework. Their work distinctly differs from ours, as they focus on the task of text completion, providing a short starting context to the model. In contrast, our work tackles the more complex task of text generation with limited input context and no specific starting point for the model.

\section{Conclusion}
In this work, we propose the first benchmark for personalized long-text generation called LongLaMP. Nearly all applications involving language generation could potentially benefit from personalized long-text generation tailored to individual users or contexts.
We investigate a retrieval-augmented generation framework, experimenting with different LLMs across diverse settings, including both fine-tuning and zero-shot settings.
Additionally, we investigate the impact of employing different retrieval methods and varying the number of documents retrieved from user profiles. 
Overall we see an average improvement of 30.21\% with ROUGE-1 metric and 47.5 \% with ROUGE-L across all tasks. 
These findings demonstrate the importance of personalization for the majority of applications involving long-text generation.
The proposed benchmark and findings pave the way for further research into personalized long-text generation, which has wide-ranging implications for enhancing user experiences and tailoring language generation to specific individuals.

%\newpage
\section{Limitations} % REQUIRED
To combat potential pre-training exposure and introduce a novel element, we use the Avocado dataset for Personalizing Email Completion into the LongLaMP benchmark. Unlike other tasks using publicly available datasets, this is a private dataset, allowing our models to engage with fresh, previously unseen data and enhancing the rigor of our evaluations.
The standard evaluation metrics used have limitations in fully capturing the complexities of long text generation and personalization. However, these metrics are widely recognized and provide a consistent framework for comparing model performance, serving as useful indicators of textual alignment and overlap, which are valuable in many applications.
Fine-tuning language models on user data raises privacy risks, including data exposure through memorization of training data \cite{carlini2019secret} and inference attacks exploiting model outputs to reveal sensitive user information \cite{shokri2017membership}. While these issues are critical, this paper does not delve into the resolution of these privacy concerns within the context of personalization.

\section{Ethics Statement}  % REQUIRED
Our paper introducing a benchmark for personalized long-text generation acknowledges the potential ethical implications inherent in using large language models for such applications.
We have taken comprehensive steps to ensure that our research adheres to the highest ethical standards, particularly concerning data privacy and the responsible use of AI.
The Avocado Research Email collection is meticulously managed under a stringent confidentiality agreement ensuring secure maintainence and limited authorized access to the dataset guaranteeing that it remains completely inaccessible to the public. 
This ethics statement reflects our dedication to conducting responsible research and our commitment to advancing the field of AI in a manner that respects individual privacy rights and promotes the ethical use of technology.

\newpage
\bibliography{main}

\appendix

\section*{Appendix}
\label{sec:appendix}

\section{Benchmark Details}
\label{appendix: data-creation-details}
This section provides details on the LongLaMP benchmark.
Table \ref{tab:task-stats} contains statistics on each task in LongLaMP and Table \ref{tab:summary} describes the differences and novelty of each task.

\subsection{Personalized Email Completion}
\label{appendix:personalized-email-generation-appendix}
Email completion is a task that can significantly benefit from personalization \cite{trajanovski-etal-2021-text}. In this task, we require the language models to complete the email, $y$, based on the given input, $x$ comprising of the subject of the email, a part of the email, and subject-email pairs previously authored by the user captured as the user profile, $P_u$. To create this task, we utilized the private email collection dataset known as the Avocado Research Email Collection \cite{Oard2015Avocado}.

\textbf{Data Curation:} The initial step in curating this task involved filtering out emails with subject lengths under five words and content under 64 words, to maintain a substantive informational base for text generation and adherence to long-form criteria. The remaining emails were then organized by sender's email address, selecting only those with a sending frequency of 10 to 200 emails, aligning with established methodologies \cite{Salemi2023LaMPWL}. Despite the dataset's limited size of 279 users, extensive measures were implemented to ensure data sufficiency, as detailed in subsequent sections. For illustrative purposes, Figure \ref{benchmark-data-email-completion} displays a constructed example. Note that the content shown is synthetically created to preserve confidentiality.

\begin{figure}[h!]
\centering
\begin{tiny}
\begin{minted}[frame=single,
               framesep=3mm,
               linenos=true,
               xleftmargin=21pt,
               breaklines=true
               tabsize=8]{json}
{
        "id": "",
        "input": "Complete the text of the following email. title: ... text: ...",
        "output": "Xxx and I thought that to complete the documentation template development efforts for phase-I we should also work on naming convention for various EP-XML elements.  I have attached a draft document for your comments.  While developing the convention we have tried to follow the convention in the popular programming languages and at the same time not have too many variations from element-to-element as that would become confusing.\n\nLooking forward to your comments.\n\nThanks,\n\nXxx",
        "profile": [
            {
                "text": "...",
                "id": "...",
                "date": "...",
                "title": "..."
            },
            ...
            {
                "text": "...",
                "id": "...",
                "date": "...",
                "title": "..."
            },
            ...
        ]
    },
    {
    "id": "",
    "input": "Complete the text of the following email. title: ... text: ...",
    "output": "...",
    "profile": [
        {
            "text": "...",
            "id": "...",
            "date": "...",
            "title": "..."
        },
        ...
        {
            "text": "...",
            "id": "...",
            "date": "...",
            "title": "..."
        },
        ...
    ]
    }
\end{minted}
\end{tiny}
\caption{Personalized email completion task schema. The $input$ represents the input prompt containing the title and part of the email. The $output$ represents the email content. The $profile$ section captures previous user-authored emails.}
\label{benchmark-data-email-completion}
\end{figure}

\textbf{User Setting:}
The users are divided into training, validation, and test sets in an 75\%, 15\%, 15\% split to ensure no overlap of users across the sets, thus enabling the model to generalize effectively to new, unseen users. After dividing the users, we increase the volume by randomly sampling 50\% of the emails from each user for inclusion in the respective sets. The remaining 50\% are aggregated to enrich the user profiles. After creating the individual splits, the input, $x$, was constructed by combining the subject with a prefix randomly selected from 20\% to 30\% of the email content. This led to a total of 3,286 training cases, 958 validation cases, and 823 test cases. The average input length is $46.45 \pm 21.45$, while the average output length is $92.59 \pm 60.68$. For more detailed statistics, please see the table referenced as Table \ref{tab:task-stats}.

\textbf{Temporal Setting:}
After filtering, to create the temporal setting, emails for each user were sorted chronologically. The most recent 10\% of these emails were allocated to the test set, the subsequent 10\% to the validation set, and the following 20\% to the training set. The remaining 60\% were used to enhance the user profile set. This approach is different from the temporal setting construction of other tasks in the LongLaMP Benchmark because of the size of the source dataset. This approach increases the dataset's volume and also ensures that each segment uniquely represents different temporal phases of the user’s email activity, facilitating a realistic and thorough evaluation of the model’s performance over time. After assembling the initial sets, a random selection from the validation and test sets is made to finalize the datasets for the task. After creating the individual splits, the input, $x$, was constructed by combining the subject with a prefix randomly selected from 20\% to 30\% of the email content. This led to a total of 3234 training cases, 833 validation cases, and 818 test cases. The average input length is $46.75 \pm 21.94$, while the average output length is $92.80 \pm 62.69$. For more detailed statistics, please see the table referenced as Table \ref{tab:task-stats}.

\textbf{Discussion \& Challenges:} 
The Avocado Research Email Collection is distinctive within our benchmark as it is a private dataset, likely excluded from the pre-training of the models used in our work. This exclusivity presents a unique challenge, offering a rigorous test of the models' personalization capabilities with entirely novel data. Furthermore, the email tone can fluctuate drastically for a single user based on situational factors. The same individual may employ a formal, professional tone when corresponding with colleagues or supervisors, yet switch to a more casual, friendly style when emailing close friends or family members. The inherently variable nature of email writing provides a testbed in assessing adaptability and personalization in language models.

\subsection{Personalized Abstract Generation}
\label{appendix:personalized-abstract-generation-appendix}
Each researcher has a unique writing style, characterized by factors such as the structure of their arguments, the use of domain-specific language, and the tone they employ. This writing style can be heavily influenced by the research field, the intended audience, the specific conference or publication venue. To test the scenarios where the generated output requires a domain specific knowledge for expert audience we curate the Personalized Abstract Generation as one of the benchmarks for LongLaMP. The expected output of this task is a scientific abstract conditioned on an input consisting of the title of paper and selected keywords of the abstract. An example can be seen in the Figure \ref{benchmark-data-example-abstraction}.

\textbf{Data Curation:}
To generate the data samples, we leverage the Citation Network Dataset (V14) \cite{Tang2008ArnetMinerEA}, which comprises 5,259,858 papers and 29 features per paper. From this dataset, we only utilize the following features: \textit{id}, \textit{title}, \textit{abstract}, \textit{authors}, \textit{year}, and \textit{language}. Furthermore, we filter out any papers that are not written in English or are missing the abstract, title, or year. The next stage is obtained by grouping the remaining data by author name and only considering authors (i.e., data points) that have at least 70 publications, ensuring a sufficient amount of data for each author for further experimentation. For each author, one of their publications is selected as the input, chosen randomly for the user setting, or chronologically for the temporal setting (as described in detail below). The title of this selected publication serves as the input, while the abstract is treated as the target output. Since the current input is only the title of the paper, which may not provide sufficient information about the methodology or main contributions needed to generate an informative abstract, keywords  are extracted from the output (i.e., the abstract) for each data point. This is accomplished by using the following prompt:

\begin{formal}
\small
\textit{\tt Mention 5 short keywords of the following abstract of the paper that shows their main findings and claims:}\\
\centerline{\textit{\tt [Abstract]}}
\textit{\tt[OUTPUT] Keyword 1, Keyword 2 , Keyword 3 , Keyword 4, Keyword 5 }
\textit{\tt \\}\\
\end{formal}
\noindent

The extracted keywords are then appended to the input title, along with task-specific instructions, resulting in each data point having following the format depicted in Figure~\ref{benchmark-data-example-abstraction}.  Providing this additional context aids the model in generating a comprehensive abstract. This step ensures that the input contains relevant details about the paper's content and contributions, facilitating more accurate and complete generation. Furthermore, for every user, we have a set of documents, that include the id, title of the paper, abstract, and year it was published, which constitutes the profile.

\textbf{User Setting:}
To create the user setting, the data is randomly split into train,validation , and test sets (60\%, 20\%, 20\%). For each author, one of their publications is randomly selected. The title of this publication serves as the input, while the abstract is treated as the target output. The remaining publications for that author comprise the profile. There are no shared user's among splits. This resulted in a total of 13696 train cases , 3454 validation cases and 3454 test cases. The input length is average of $20.090 \pm 4.53$ and the average output length is $143.722 \pm 62.911$. More detailed statistics can be referenced in the Table \ref{tab:task-stats}.

\textbf{Temporal Setting:}
For the temporal setting, the data is split by first identifying the three most recent publications for each author. The most recent publication is assigned to the test set, the second most recent to the validation set, and the third most recent to the training set. All remaining publications for that author comprise their profile. The validation and test sets are then randomly subsampled to maintain 20\% of the data points in each. For every data point across the training, validation, and test sets, the title of the corresponding publication serves as the input, while the abstract is treated as the target output to be generated. This temporal setting allows for evaluating the model's ability to adapt to an author's evolving writing style over time. There will be shared authors among the splits. This resulted in a total of 17270 train cases , 3454 validation cases and 3454 test cases. The input length is average of $34.638 \pm 6.003$ and the average output length is $162.485 \pm 70.020$. More detailed statistics can be referenced in the Table \ref{tab:task-stats}.

\begin{figure}[h!]
\begin{tiny}
\begin{minted}[frame=single,
               framesep=3mm,
               linenos=true,
               xleftmargin=21pt,
               breaklines=true
               tabsize=8]{json}
{
  "name": "Roberto Battiti", 
  "input": "Generate an abstract for the title \"Reactive and dynamic local search for max-clique: Engineering effective building blocks\" using the following items: 1. Algorithmic building blocks \n2. Maximum clique problem \n3. Dynamic local search \n4. Long-term memory \n5. Implementation and scalability.",   
  "output": "This paper presents the results of an ongoing investigation about how different algorithmic building blocks contribute to solving the maximum clique problem. We consider greedy constructions, plateau searches, and more complex schemes based on dynamic penalties and/or prohibitions, in particular the recently proposed technique of dynamic local search and the previously proposed reactive local search (RLS). We design a variation of the original RLS algorithm where the role of long-term memory (LTM) is increased (RLS-LTM). In addition, we consider in detail the effect of the low-level implementation choices on the CPU time per iteration. We present experimental results on randomly generated graphs with different statistical properties, showing the crucial effects of the implementation, the robustness of different techniques, and their empirical scalability.",
  "profile": [
        {
            "id": "...",
            "title": "...",
            "abstract": "....",
            "year": ...
        },
        ...,
        {
            "id": "...",
            "title": "...",
            "abstract": "....",
            "year": ...
        },
        ...
    ] 
},

...,

{
  "name": "...", 
  "input": "...", 
  "output": "...",
  "profile": [
        {
            "id": "...",
            "title": "...",
            "abstract": "....",
            "year": ...
        },
        ...,
        {
            "id": "...",
            "title": "...",
            "abstract": "....",
            "year": ...
        },
        ...
  ] 
}, 

...
\end{minted}
\end{tiny}
\caption{%
Personalized abstract generation task schema. 
Note that input is the prompt for the generation question for the user, and output is the ground-truth generation for that specific user's input question.
Further, profile (\eg, set of text documents and profile information for that user) is a (possibly) large set of text documents used by our retrieval model for generating personalized abstracts.
}
\label{benchmark-data-example-abstraction}
\end{figure}

\textbf{Discussion \& Challenges:} 
A typical scientific abstract spans multiple paragraphs and requires precise, coherent, and objective language to distill complex ideas, methods, and findings in an unbiased manner. Furthermore, scientific abstracts necessitate the incorporation of domain-specific information, a task that large language models (LLMs) often struggle with \cite{bang-etal-2023-multitask}. The combination of these factors – the need to capture individual writing styles, maintain coherence and objectivity across multiple paragraphs, and accurately incorporate domain-specific information – presents a challenging testbed for personalized long text generation tasks. Successful approaches to this challenge could pave the way for more sophisticated personalized text generation systems capable of producing high-quality, tailored content for academic and research settings.

\subsection{Personalized Review Writing}
\label{appendix:personalized-product-generation-appendix}
Each consumer review reflects the unique perspective and expectations of the reviewer about a product, heavily influenced by personal experiences and specific product features. The style and content of these reviews are adapted to cater to a broad audience of potential buyers. To precisely assess the ability of models to generate tailored and authentic content, we've established the Personalized Review Writing as one of the benchmarks for LongLaMP. This task is crafted to assess the model's capability to generate a product review, denoted as $y$, from input $x$ and user profile $P_u$. The input $x$ encompasses the product description, the user's product rating, and a summary of the user's experience. The user profile $P_u$ consists of other reviews made by the user described using the review text, summary of the review, rating given by the user and description of the product. An example data entry is provided in Figure~\ref{benchmark-data-example-product-review-writinf}.
\textbf{Data Curation:}
To generate data samples, we leverage the Amazon Reviews Dataset \cite{Ni2019JustifyingRU}, which comprises 150 million reviews and 12 features per review. 
The Amazon dataset structure can be seen in Figure \ref{amazon-dataset-structure}.
We also utilize a separate dataset to retrieve the metadata related to every product whose structure we can see in Figure \ref{amazon-metadata-dataset-structure}.
\\From the Amazon Product Reviews dataset, we only utilize the following features : overall, reviewerID, reviewText, summary, asin, reviewTime. \textit{overall} refers to the rating given by the user for the product. \textit{reviewerID} refers to the ID of the user/reviewer. \textit{reviewText} refers to the actual text generated by the user. \textit{summary} depicts the summary of the user's review about the product. \textit{asin} refers to the ID of the product reviewed by the user. \textit{reviewTime} depicts the time at which the review was published. 

\begin{figure}[ht!]
\begin{tiny}
\begin{minted}[frame=single,
               framesep=3mm,
               linenos=true,
               xleftmargin=21pt,
               breaklines=true
               tabsize=8]{vim}
{
  "image": ["https://images-na.images-amazon.com/71eG.jpg"], 
  "overall": 5.0, 
  "vote": "2", 
  "verified": True, 
  "reviewTime": "01 1, 2018", 
  "reviewerID": "AUI6WTTT0QZYS", 
  "asin": "5120053084", 
  "style": {
    "Size:": "Small", 
    "Color:": "Pink"
    }, 
  "reviewerName": "Abbey", 
  "reviewText": "I now have 4 of the 5 available colors of this tutu... ", 
  "summary": "Comfy, flattering, discreet--highly recommended!", 
  "unixReviewTime": 1514764800
}
\end{minted}
\end{tiny}
\caption{Structure of the Amazon Product Review dataset} 
\label{amazon-dataset-structure}
\end{figure}

\begin{figure}[ht!]
\begin{tiny}
\begin{minted}[frame=single,
               framesep=3mm,
               linenos=true,
               xleftmargin=21pt,
               breaklines=true
               tabsize=8]{vim}
{
  "asin": "5120053084",
  "title": "Girls Ballet Tutu Zebra Hot Pink",
  "feature": ["Botiquecutie Trademark exclusive Brand",
              "Hot Pink Layered Zebra Print Tutu",
              "Fits girls up to a size 4T",
              "Hand wash / Line Dry",
              "Includes a Botiquecutie TM Exclusive hair flower bow"],
  "description": "This tutu is great for dress up play for your little ballerina. Botiquecute Trade Mark exclusive brand. Hot Pink Zebra print tutu.", 
  "price": 3.17,
  "imageURL": "http://ecx.images-amazon.com/images/I/51fAm.jpg",
  "imageURLHighRes": "http://ecx.images-amazon.com/images/I/51fAmVkTbyL.jpg",
  "also_buy": ["B00JHONN1S", "B002BZX8Z6"],
  "salesRank": {"Toys & Games": 211836},
  "brand": "Coxlures",
  "categories": [["Sports & Outdoors", "Other Sports", "Dance"]]
}
\end{minted}
\end{tiny}
\caption{Structure of the Amazon Product Review Metadata dataset} 
\label{amazon-metadata-dataset-structure}
\end{figure}

The data undergo a detailed filtering process to create the LongLaMP benchmark. This ensures that the selected data entries are complete in essential fields, have sufficient length in the review text, and come from users with a substantial review history. The filtering process is performed as follows.
Each \textit{reviewText} must surpass its corresponding summary in length and should contain no fewer than 120 words to ensure the long form criteria. Each \textit{summary} must have at least 4 words, providing a basic yet sufficient understanding of the review's content. Only reviews from users with at least 50 contributions are considered, such that there is enough data to provide context for personalization. This resulted in 25318 users with an average number of reviews of 151 and an average output length of 336.84. All the reviews are aggregated based on the reviewerID. Furthermore, for every review, the \textit{asin} field is used to retrieve the product description of the reviewed product from the Metadata dataset as given in its \textit{description} field. This description is integrated into the filtered dataset, thus adding further contextual depth for generating the review text. Any reviews that do not contain a corresponding product description are removed from the dataset. It should be noted that many product descriptions contained some links and other text under HTML tags. To make sure that the product descriptions are intelligible, any HTML tags are removed from each of the descriptions. This process of filtering resulted in 24552 users with an average number of reviews of 51.71 and an average output length of 320.42. The user and temporal setting proceed to divide the filtered dataset into training, validation, and test segments using selection processes that are explained in the next sections.
From the training, validation, and testing datasets obtained from user and temporal settings, there was a very small subset of users that had less than 4 reviews in their profile who were removed from the segments to make sure the retrieval process contained enough documents to retrieve from.
Note that in Figure~\ref{benchmark-data-example-product-review-writinf}, we show the id of the reviewer, the input prompt $\vx$, and the ground-truth long-text $\vy$ output that we aim to generate.
Furthermore, for every user, we have a set of documents, that include the \textit{overall}, \textit{reviewText}, \textit{summary}, \textit{reviewTime} and \textit{description}.

\textbf{User Setting:} 
After implementing the outlined filtering and aggregation procedures, the resulting intermediate dataset comprises posts aggregated by the reviewer. From this collection, a single product review made by a user is randomly chosen for inclusion in the dataset, while the remaining reviews from the same user constitute the user profile section. Following the selection of this product review, the input field $x$ is formulated utilizing the rating, product description, and summary. The output field $y$ is formulated using the actual review text. The dataset entry is created using the reviewerId, input, output, and profile. The final dataset consisting of all the entries is divided into training, testing, and validation subsets, adhering to an 80-10-10 percentage split, respectively. After filtering out users with less than 4 reviews in their profile, there was a total of 14745 train cases, 1826 validation cases, and 1822 test cases. The average input length is 119.39 $\pm$ 73.06, while the average output length is 304.54 $\pm$ 228.61. For more detailed statistics, please refer to Table \ref{tab:task-stats}.

\textbf{Temporal Setting:} 
Similarly, after the same initial filtering, a temporal setting is generated by ordering the posts chronologically by the 'reviewTime' field in the Amazon Product Review dataset depicted in Figure \ref{amazon-dataset-structure}. The most recent review is allocated to the test set, the second most recent to the validation set, and the third to the training set. The remaining reviews are aggregated into the profile section. Additionally, a subset of 10\% from the test and validation sets is utilized to create the final test and validation sets. After filtering out users with less than 4 reviews in their profile, there was a total of 16197 train cases, 1831 validation cases, and 1784 test cases. The average input length is 121.68 $\pm$ 71.63, while the average output length is 296.15 $\pm$ 229.13. For more detailed statistics, please refer to Table \ref{tab:task-stats}.

\textbf{Discussion \& Challenges:} 
The Amazon Product Review dataset is known for its large volume of reviews covering a wide range of products. This variety allows for training and/or testing LLMs across diverse domains, which can be essential for assessing how well a model can personalize responses based on different user preferences or product categories. The reviews are user-generated, offering authentic insights into consumer preferences, sentiments, and language use. This natural language data can help in prompting and/or fine-tuning LLMs to understand and generate human-like text, which is crucial for personalization. Along with reviews, the Metadata dataset typically includes additional information such as product descriptions which can be instrumental in developing sophisticated personalization features. The data reflects real-world purchasing and review-writing behaviors, which are relevant for applications in e-commerce, recommendation systems, and targeted advertising. Understanding these patterns can enhance the ability of LLMs to provide personalized content that aligns with user interests and past behaviors. 
\\The key difference between the Review Writing dataset and the others in the benchmark is that it focuses on consumer opinions and experiences with specific products. The dataset provides a rich corpus of varied user feedback which can be leveraged to train models to produce detailed and nuanced text that mimics authentic customer evaluations. The language is typically subjective, based on personal experience, and directed toward product features, quality, and user satisfaction. The language complexity of the reviews in the dataset will vary from others in the benchmark, with mostly the reviews being straightforward and informal and aimed at describing personal user experiences.

\begin{figure}[ht!]
\begin{tiny}
\begin{minted}[frame=single,
               framesep=3mm,
               linenos=true,
               xleftmargin=21pt,
               breaklines=true
               tabsize=8]{json}
{
  "reviewerId": "A1KSRHAXD67HI0", 
  "input": "Generate the review text...", 
  "output": "Alpha Goddess had so much potential!  I was totally sucked into the idea of the story and that gorgeous cover.  Im don't know a lot about Hindu mythology, but I find the little I have read so interesting.  I was really excited to read a story that brought Indian gods and goddesses to present life.  Unfortunately there was just way too much going on.  I could barely keep all the parties straight and at one point I realized I just didn't care to try anymore.\n\nI liked Sera in the beginning.  Shes having terrifying dreams as she turned 16-bloody kisses and monsters and is hiding them from her parents. I won't even get into the love triangle she gets herself into-because shes not just Sera anymore now. Shes now remembering her past lives and loves-and they're still part of her life.  It was all too much to keep straight.\n\nThe end explodes into a major battle between gods and demons, but even that couldn't keep my attention.  Sera stopped to chat way too many times it seemed when she could have been saving the world.  She was way too consumed with her love options.  The concept of this book was fantastic, but the execution was really lacking.",
  "profile":[
        {
                "overall": "...",
                "reviewText": "...",
                "summary": "...",
                "description": "..."
        },
        ...,
        {
                "overall": "...",
                "reviewText": "...",
                "summary": "...",
                "description": "..."
        },
        ...
    ] 
},
...
{
  "reviewerId": "...", 
  "input": "...", 
  "output": "...",
  "profile":[
        {
                "overall": "...",
                "reviewText": "...",
                "summary": "...",
                "description": "..."
        },
        ...,
        {
                "overall": "...",
                "reviewText": "...",
                "summary": "...",
                "description": "..."
        },
        ...
    ] 
}
\end{minted}
\end{tiny}
\caption{Personalized Review Writing Task Schema. Note that the \textit{input} represents the prompt that contains the user's rating, summary of the review, and product description. The \textit{output} represents the ground-truth review text generated by the user. The \textit{profile} section captures the reviews previously made by the user.} 
\label{benchmark-data-example-product-review-writinf}
\end{figure}

\subsection{Personalized Topic Writing} 
\label{appendix:personalized-topic-generation-appendix}

The dataset is created from the Reddit TL;DR dataset \cite{Vlske2017TLDRMR}. The task involves generating the content of a Reddit post, $y$, based on the post's summary, $x$, and the user's previous posts, $P_u$. Figure \ref{benchmark-data-topic-generation} displays a sample data point for reference. The user profile, $P_u$ is a compilation of summary-content pairs authored by the user previously. This dataset is curated by capturing posts and comments from a broad range of subreddits thereby capturing discussions across a variety of topics. TL;DR is a community-encouraged practice and thereby all the summaries are user-generated, the dataset filters out any bot-generated posts making it a suitable test bed for personalization.

\begin{figure}[h!]
\centering
\begin{tiny}
\begin{minted}[frame=single,
               framesep=3mm,
               linenos=true,
               xleftmargin=21pt,
               breaklines=true
               tabsize=8]{json}
{
        "author": "ENovi",
        "input": "Generate content for the reddit post .... .",
        "output": "I do too. When Pujols was suffering from plantar fasciitis he struggled terribly, leading everyone to declare that he was a total has-been. In 2012 (after a terrible April which people seem to think represented his entire year) he hit .285/.343/.516 with 30 bombs and over 100 RBIS. The next year, when he was injured, he hit only .258/.330/.437 with only 17 home runs in limited playing time. This was due to plantar fasciitis. \n When Pujols came back in '14 he hit .272/.324/.466 with 28 home runs and another 105 RBIs. The numbers weren't mind blowing (and a bit low for a guy like Albert) ....",
        "profile": [
            {
                "author": "ENovi",
                "content": "...",
                "summary": "..."
            },
            ...
            {
                "author": "ENovi",
                "content": "...",
                "summary": "..."
            },
            ...
        ]
    },
    ...
    {
    "author": "...",
    "input": "...",
    "output": "...",
    "profile": [
        {
            "author": "...",
            "content": "...",
            "summary": "...",
        },
        ...
        {
            "author": "...",
            "content": "...",
            "summary": "...",
        },
        ...
    ]
}
\end{minted}
\end{tiny}
\caption{Personalized topic writing task schema. The $input$ represents the summary input prompt containing the summary of the post. The $output$ represents the content of the post. The $profile$ section captures posts previously made by the user containing both the content and summary.}
\label{benchmark-data-topic-generation}
\end{figure}

\textbf{Data Curation:} 
Derived from an initial corpus comprising roughly four million entries, this dataset was curated using a set of filtering criteria that is focused on the fields: post summaries, content, authors, and identifiers. 

The summary of each post serves as the input $x$, while the content is designated as the output $y$. The dataset is created by applying a filtering criterion aimed at ensuring ample profile data for each participant and guaranteeing that the text generated adheres to long-form criteria. This involves several layers of filtering: initially, the length of a post's content must exceed that of its summary. Furthermore, the content is required to contain a minimum of 50 words to align with the objective of generating long text. Additionally, only users who have made a minimum contribution of 16 posts are considered for inclusion. This threshold is determined by analyzing the distribution curve of users versus posts to ensure a robust dataset for both training and evaluation purposes. Figure \ref{benchmark-data-topic-generation} presents a sample data point of this dataset.

\textbf{User Setting:} 
Subsequent to the application of these filters, a random selection process is employed to choose one post per user for inclusion in the dataset, while aggregating the rest of the user's contributions to construct comprehensive user profiles. To maintain integrity and prevent data leakage, we meticulously divide the dataset into training, validation, and test segments, ensuring each set comprises a unique set of users. The division follows a 70-15-15 ratio for the training, validation, and testing segments, respectively. Table \ref{tab:task-stats} shows the statistics of the created dataset.

\textbf{Temporal Setting:} 
After applying the filtration criteria, given the absence of temporal data within the Reddit dataset, for each author, one post is randomly assigned to each of the training, testing, and validation sets. The posts not selected for these sets are then consolidated into the profile section, ensuring consistency across the dataset divisions. Following this distribution, a subset amounting to 15\% of the initial dataset size is randomly designated to form the testing and validation sets, thereby facilitating a structured yet randomized methodology in constructing the dataset. Table \ref{tab:task-stats} provides a detailed statistical breakdown of the dataset thus configured. 

\textbf{Discussion \& Challenges:}
The Reddit TL;DR dataset comprises a diverse collection of posts sourced from various subreddits, each presenting discussions that span a broad spectrum of topics and exhibit varied writing styles, even among entries by the same author. This variability presents a challenge for the development of personalized generation systems, which must adapt to differing content and stylistic nuances based on the discussion topic. In addition to the common challenges associated with long-text generation, such as contextual understanding, the generation of plausible details, and creative gap-filling, this dataset introduces a unique complexity. The summaries within this dataset are manually created by users, not derived through automated means. This human-driven process could capture intricate linguistic nuances and variability in summarization styles that automated techniques might fail to replicate, thereby providing an excellent platform for evaluating personalized long-text generation methods.

\begin{table*}[htbp]
\centering
\resizebox{\textwidth}{!}{%
\begin{tabular}{r cccc}
\toprule

& \textbf{Personalized} & \textbf{Personalized} & \textbf{Personalized} & \textbf{Personalized}\\
 
& \textbf{\textit{Abstract Generation}} & \textbf{\textit{Topic Writing}} & \textbf{\textit{Review Writing}} & \textbf{\textit{Email Writing}} \\ 
\midrule
\textbf{Audience} & Researchers & General/Niche & Consumers & Anyone \\
\textbf{Purpose} & Elaborate, Educate & Inform, Discuss & Provide Insights, Opinions & Communicate, Interact \\

\textbf{Style} & Formal, Precise & Any & Informal/Semi-formal & Any \\

\textbf{Content} & Methodology, Results & Posts, Discussions, Comments & Product Features, Ratings, Feedback & Email body, Subject

\\
\textbf{Credibility/Trust} & Authoritative Sources & Author's Authenticity & Reviewer's Authenticity & Sender's Reputation \\

\textbf{Length} & Fixed (Word Limit) & Moderate Length & Variable & Variable \\
\textbf{Structure} & Title, Abstract & Review text, Rating, Summary, Metadata & Post content, summary & Email title, text \\
\textbf{Use Case} & Academic, Research & E-commerce, Marketing & Social Media Analytics & Customer Support, Business Communication \\
\bottomrule
\end{tabular}
}
\caption{Summary of Differences and Novelty in Text Generation Tasks}
\label{tab:summary}
\end{table*}

\begin{figure}[t!]
\centering
\includegraphics[width=1.00\linewidth]{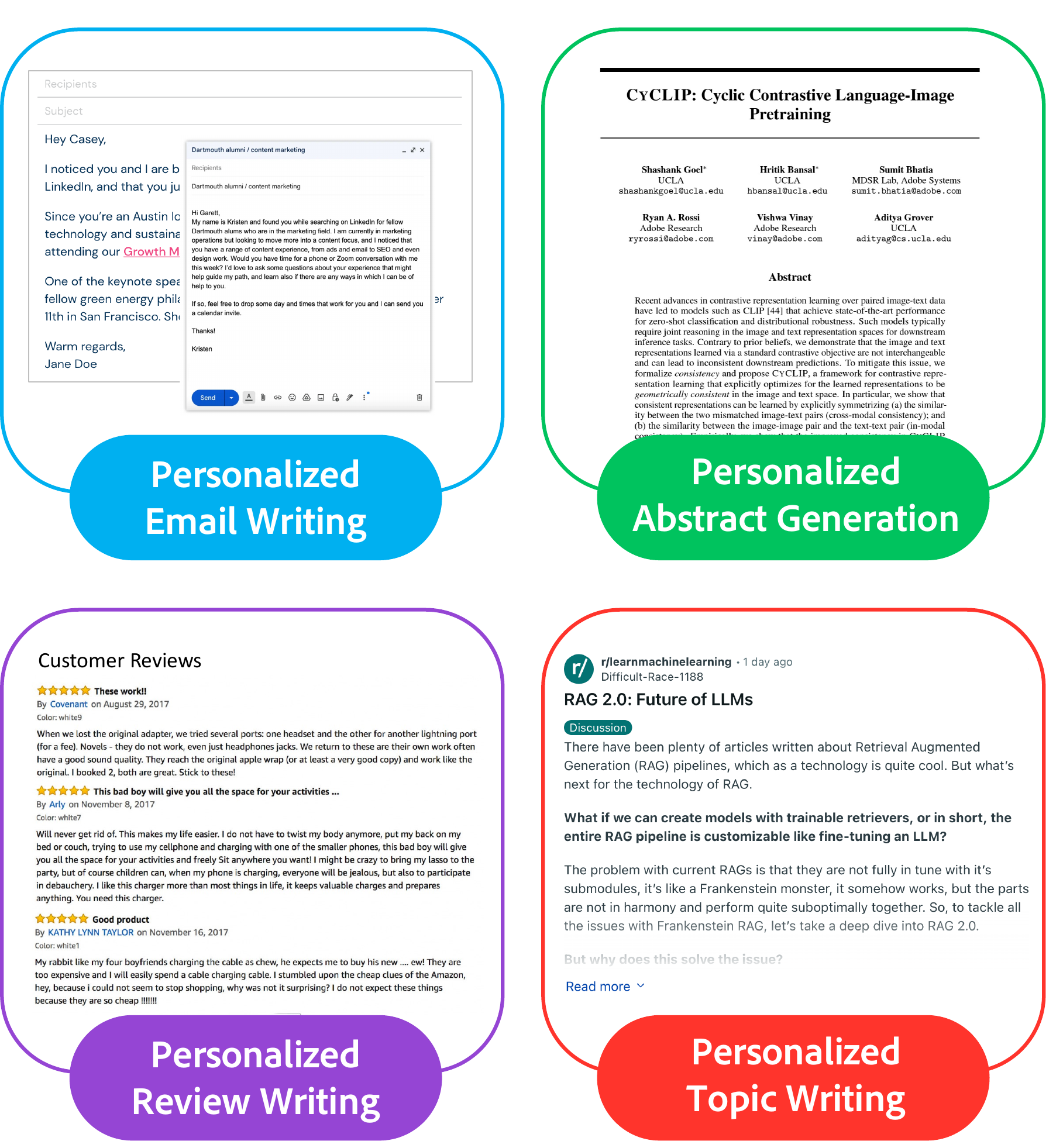}
\caption{%
Overview of the \underline{Long}-text \underline{L}anguage \underline{M}odel \underline{P}ersonalization (LongLaMP) Benchmark. 
}
\label{fig:overview-tasks}
\end{figure}

\section{Prompts Used for Adding User Profile to the Language Model's Input}
We use one or more entries from the user profile in order to personalize the language model's input. For this purpose, we construct prompts specific to each task using the templates given in Table \ref{tab:prompts-template} which mentions how the prompts were created to perform zero-shot experiments with GPT-3.5 and fine-tuning experiments with FlanT5-base models. The prompt creation is performed in 2 steps: 1) Per Profile Entry Prompt (PPEP) creation and 2) Aggregated Input Prompt (AIP) creation. In the first step, we follow the instructions given in Table \ref{tab:prompts-template} to create a prompt for each profile entry. In the second step, we follow the instructions given in Table \ref{tab:prompts-template} to combine the PPEP prompts along with the input to be fed into the language model. In the case of the Personalized Abstract Generation task, we noticed that selecting only the first 750 words of the abstract from every profile entry helped achieve better results compared to utilizing the entire abstract. This might be due to the noise that was being added when profile entries were large. The only difference in the prompt template for the Llama-7B model is with respect to the Personalized Review Writing task in which only the first 100 words of the review text from every profile entry is used.

\begin{table*}[!t]
\centering
\resizebox{\textwidth}{!}{
\begin{tabular}{llccc c}
\toprule
{\bf Task} & {\bf Per Profile Entry Prompt (PPEP)} & {\bf Aggregated Input Prompt (AIP)}
\\
\midrule
\multirow{3}{*}{Personalized Abstract Generation} & 
extract\_first\_750\_words([ABSTRACT]) & concat([PPEP($P_1$), ... , PPEP($P_n$)],    \\
& is the abstract for the title [TITLE] & "Use the above title and abstracts as context to \\
& & understand the style and language of the user and", [INPUT])\\
\midrule
\multirow{4}{*}{Personalized Review Writing} & 
[OVERALL] is a rating for the product with description  & concat([PPEP($P_1$),... ,PPEP($P_n$)],    \\
& [DESCRIPTION]. [SUMMARY] is summary for [REVIEW TEXT] & "Following the given patterns,", [INPUT]) \\
& & \\
\midrule
\multirow{2}{*}{Personalized Topic Writing} & 
[SUMMARY] is a summary for [CONTENT] & concat([PPEP($P_1$), ... , PPEP($P_n$)],    \\
& & "Following the given patterns,", [INPUT]) \\
\midrule
\multirow{2}{*}{Personalized Email Generation} & 
[TITLE] is the title for [TEXT] & concat([PPEP($P_1$), ... , PPEP($P_n$)],    \\
& & "Following the given patterns,", [INPUT]) \\
\bottomrule
\end{tabular}}
\caption{Prompts template used to augment the input of the LM with the user profile. concat is a function that concatenates the strings in its first argument by placing the string in the second argument between them. PPEP is a function that creates the prompt for each entry in the retrieved profile entries. [INPUT] is the task’s input.}
\label{tab:prompts-template}
\end{table*}

\begin{figure}[h!]
\centering
% \hspace{-10mm}
\includegraphics[width=1.00\linewidth]{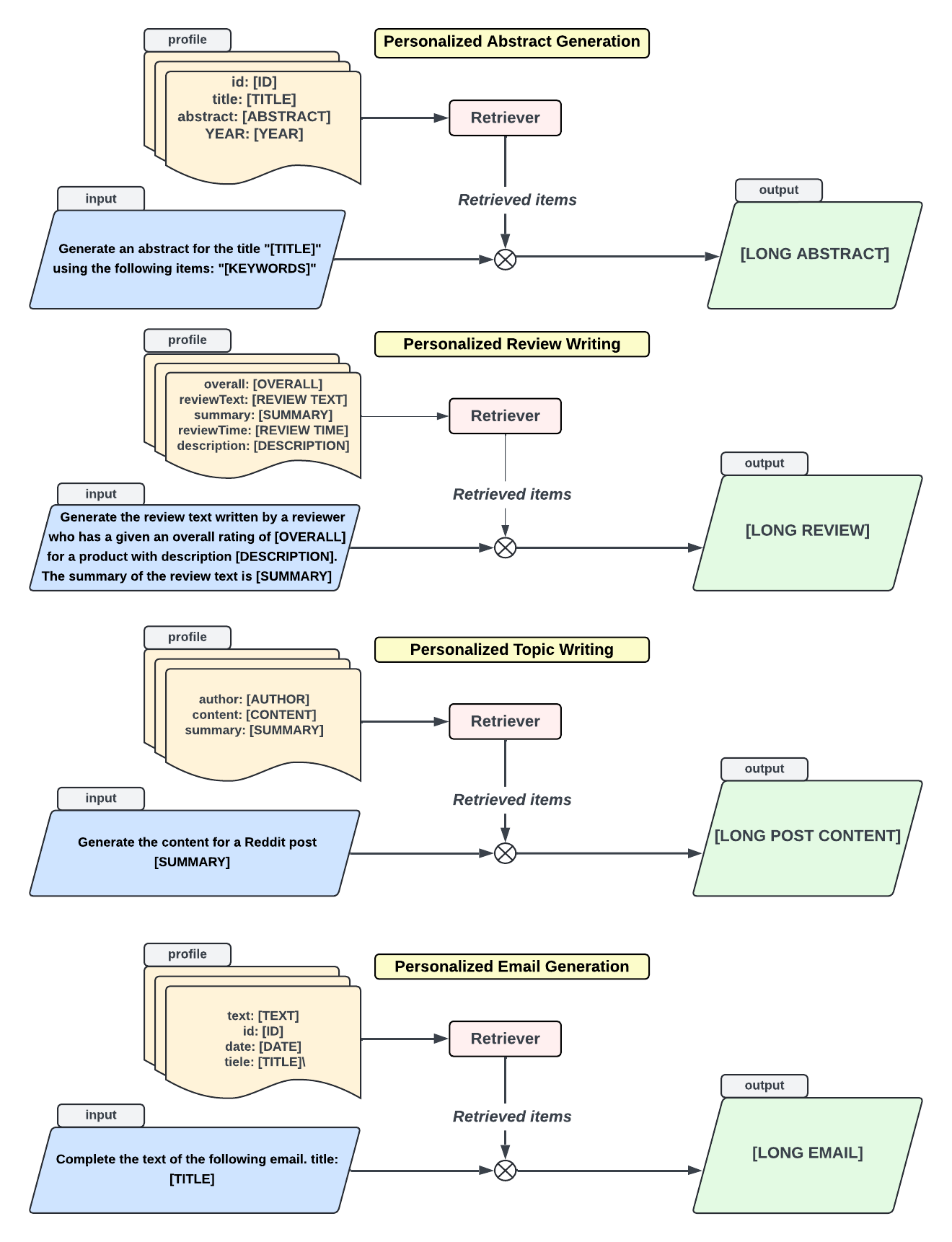}
\caption{Overview of the templates used to create each task in LongLaMP.}
\label{fig:overview-prompts}
\end{figure}

\section{Experimental Setup}
\label{appendix:experimental-setup-extra}
The GPT-3.5 model used for zero-shot experiments is set to have a maximum response length of 2048 tokens and a temperature of 0.6. 
\\The Llama 7B model used for zero-shot experiments is set to have a maximum response length of 4096 tokens and a temperature of 0.8. It also has a nucleus sampling parameter, '\textit{top\_p}' of 0.95. \\The FlanT5-base model \cite{chung2022scaling} leverages the AdamW \cite{loshchilov2018decoupled} optimizer with a learning rate of 5 x $10^{-5}$ and a batch size of 1. The maximum output length is set to 1024 tokens and the gradient accumulation steps is set to 4.  The generation model is trained for 20 epochs. All the experiments are conducted on a single Nvidia RTX8000 GPU with 49GB of GPU memory and 128GB of CPU memory.

\begin{table*}[h!]
\centering
\begin{adjustbox}{max width=\textwidth}
\begin{tabular}{lll c ccc ccc}
\toprule
% \textbf{Personalized Long-Text} \\
\multirow{2}{*}{\bf Personalized Long-Text} &  \multirow{2}{*}{\bf Setting} & \multirow{2}{*}{\bf Metric} & & \multicolumn{3}{c}{BM25} & \multicolumn{3}{c}{Contriever} \\
\cmidrule(r){5-7} \cmidrule(l){8-10}
{\bf Benchmark Data} & & & No Personal. & $k=1$ & $k=2$ & $k=4$ & $k=1$ & $k=2$ & $k=4$ \\
\midrule
\multirow{4}{*}{\shortstack[l]{{\bf Personalized}\\{\bf Abstract Generation}}} &
\multirow{2}{*}{User} & ROUGE-1 $\uparrow$ & 0.3879 & 0.4161 & 0.4206 & 0.4238 & 0.4187 & 0.4239 & \textbf{0.42521} \\
& & ROUGE-L $\uparrow$ & 0.2205 & 0.2406 & 0.2445 & 0.2469 & 0.2424 & 0.2462 & \textbf{0.2481} \\

\cmidrule{2-10}
& \multirow{2}{*}{Temporal} & ROUGE-1 $\uparrow$ & 0.3724 & 0.3887 & 0.3915 & 0.3933 & 0.3894 & 0.3939 & \textbf{0.3945} \\
&  & ROUGE-L $\uparrow$ & 0.2101 & 0.2198 & 0.2213 & 0.2227 & 0.2195 & 0.2222 & \textbf{0.2234} \\

\midrule

\multirow{4}{*}{\shortstack[l]{{\bf Personalized}\\{\bf Topic Writing}}} & 
\multirow{2}{*}{User} & ROUGE-1 $\uparrow$ & 0.2581 & 0.2805 & 0.2778 & 0.2706 & \textbf{0.2863} & 0.286 & 0.2704 \\
& & ROUGE-L $\uparrow$ & 0.1251 & 0.1328 & 0.1323 & 0.1298 & \textbf{0.1342} & 0.1341 &  0.1300 \\

\cmidrule{2-10}
& \multirow{2}{*}{Temporal} & ROUGE-1 $\uparrow$ & 0.2605 & 0.2801 & 0.2835 & 0.2816 & \textbf{0.2889} & 0.2865 & 0.2863 \\
&  & ROUGE-L $\uparrow$ & 0.1275 & 0.1335 & 0.1341 & 0.1348 & 0.1354 & 0.1361 & \textbf{0.1363}  \\

\midrule
\multirow{4}{*}{\shortstack[l]{{\bf Personalized}\\{\bf Review Writing}}} & 
\multirow{2}{*}{User} & ROUGE-1 $\uparrow$ & 0.2768 & 0.2904 & \textbf{0.2920} & 0.2910 & 0.2879 & 0.2925 & 0.2984 \\
&  & ROUGE-L $\uparrow$ & 0.1397 & 0.1454 &  0.1468 & \textbf{0.1481} & 0.1449 & 0.1485 & 0.1514 \\

\cmidrule{2-10}
& \multirow{2}{*}{Temporal} & ROUGE-1 $\uparrow$ & 0.2829 & 0.2949 & \textbf{0.2956} & 0.2923 &0.2908 & 0.2909 & 0.2882  \\
&  & ROUGE-L $\uparrow$ & 0.1412 & 0.1455 & \textbf{0.1467} & 0.1467 & 0.1451 & 0.1458 & 0.14516 \\

\bottomrule
\end{tabular}
\end{adjustbox}
\caption{The zero-shot personalized results using the GPT-3.5 model on the validation set for user and temporal settings. The best results are highlighted in bold.}
\label{table:zero-shot-validation-results-gpt}
\end{table*}

\begin{table*}[h!]
\centering
\begin{adjustbox}{max width=\textwidth}
\begin{tabular}{lll c ccc ccc ccc}
\toprule

\multirow{2}{*}{\bf Personalized Long-Text} &  \multirow{2}{*}{\bf Setting} & \multirow{2}{*}{\bf Metric} & & \multicolumn{3}{c}{BM25} & \multicolumn{3}{c}{Contriever}\\
\cmidrule(r){5-7} \cmidrule(l){8-10}
{\bf Benchmark Data} & & & No Personal. & $k=1$ & $k=2$ & $k=4$ & $k=1$ & $k=2$ & $k=4$ \\
\midrule
\multirow{4}{*}{\shortstack[l]{{\bf Personalized}\\{\bf Abstract Generation}}} &
\multirow{2}{*}{User} & ROUGE-1 $\uparrow$ & 0.3503 & 0.3477 & 0.3834 & 0.3907 & 0.3504 & 0.3817 & \textbf{0.3921} \\
& & ROUGE-L $\uparrow$ & 0.1914 & 0.1937 & 0.2125 & 0.2178 & 0.1956 & 0.2125 &  \textbf{0.2188} \\

\cmidrule{2-10}
& \multirow{2}{*}{Temporal} & ROUGE-1 $\uparrow$ & 0.3547 & 0.3309 & 0.3645 & 0.3712 & 0.3318 & 0.3669 & \textbf{0.3716}\\
&  & ROUGE-L $\uparrow$ & 0.1896 & 0.1800 & 0.1967 & \textbf{0.2012} & 0.1808 & 0.1973 & 0.2011\\
\midrule
\multirow{4}{*}{\shortstack[l]{{\bf Personalized}\\{\bf Topic Writing}}} & 
\multirow{2}{*}{User} & ROUGE-1 $\uparrow$ & 0.2187 & 0.2391 & 0.2364 & 0.2061 & \textbf{0.2501} & 0.2487 & 0.2291 \\
& & ROUGE-L $\uparrow$ & 0.1102 & 0.1180 & 0.1171 & 0.1021 & \textbf{0.1236} & 0.1213 & 0.1133 \\

\cmidrule{2-10}
& \multirow{2}{*}{Temporal} & ROUGE-1 $\uparrow$ & 0.2224 & 0.2348 & 0.2285 & 0.2003 & \textbf{0.2451} & 0.2406 & 0.2227 \\
&  & ROUGE-L $\uparrow$ & 0.1119 & 0.1179 & 0.1143 & 0.1003 & \textbf{0.1208} & 0.1173 & 0.1095 \\
\midrule
\multirow{4}{*}{\shortstack[l]{{\bf Personalized}\\{\bf Review Writing}}} & 
\multirow{2}{*}{User} & ROUGE-1 $\uparrow$ & 0.2694 & 0.2844 & 0.2861 & 0.2804 & 0.2868 & \textbf{0.2881} & 0.2800 \\
&  & ROUGE-L $\uparrow$ & 0.1318 & 0.1376 & 0.1367 & 0.1354 & \textbf{0.1389} & 0.1384 & 0.1355 \\
\cmidrule{2-10}
& \multirow{2}{*}{Temporal} & ROUGE-1 $\uparrow$ & 0.2731 & 0.2840 & \textbf{0.2879} & 0.2826 & 0.2836 & 0.2863 & 0.2825\\
&  & ROUGE-L $\uparrow$ & 0.1336 & 0.1366 & \textbf{0.1372} & 0.1357 & 0.1369 & 0.1372 & 0.1357\\
\midrule
\multirow{4}{*}{\shortstack[l]{{\bf Personalized}\\{\bf Email Writing}}} & 
\multirow{2}{*}{User} & ROUGE-1 $\uparrow$ & 0.1925 & 0.3579 & 0.3700 & 0.3692 & 0.3560 & 0.3766 & \textbf{0.3777} \\
&  & ROUGE-L $\uparrow$ & 0.1194 & 0.3043 & 0.3158 & 0.3219 & 0.2987 & 0.3217 & \textbf{0.3291} \\
\cmidrule{2-10}
& \multirow{2}{*}{Temporal} & ROUGE-1 $\uparrow$ & 0.1819 & 0.2737 & 0.2858 & 0.2904 & 0.2885 & 0.2889 & \textbf{0.2911}\\
&  & ROUGE-L $\uparrow$ & 0.1156 & 0.2146 & 0.2317 & 0.2364 & 0.2276 & 0.2333 & \textbf{0.2389}\\
\bottomrule
\end{tabular}
\end{adjustbox}
\caption{The zero-shot personalized results using the LlaMA-7B model on the validation set for user and temporal settings. The best results are highlighted in bold.
}
\label{table:zero-shot-validation-results-LLaMA7}
\end{table*}

\begin{table*}[h!]
\centering
\begin{adjustbox}{max width=\textwidth}
\begin{tabular}{lll c ccc ccc}
\toprule
\multirow{2}{*}{\bf Personalized Long-Text} &  \multirow{2}{*}{\bf Setting} & \multirow{2}{*}{\bf Metric} & & \multicolumn{3}{c}{BM25} & \multicolumn{3}{c}{Contriever} \\
\cmidrule(r){5-7} \cmidrule(l){8-10}
{\bf Benchmark Data} & & & Non-Personalized & $k=1$ & $k=2$ & $k=4$ & $k=1$ & $k=2$ & $k=4$ \\
\midrule
\multirow{6}{*}{\shortstack[l]{{\bf Personalized}\\{\bf Abstract Generation}}} &
\multirow{2}{*}{User} & ROUGE-1 $\uparrow$ & 0.3643 & 0.3666 & 0.3658 & 0.3194 & \textbf{0.3667} & 0.3660 & 0.3194 \\
& & ROUGE-L $\uparrow$ & 0.2427 & 0\textbf{.2438} & 0.2423 & 0.2090 & 0.2437 & 0.2423 & 0.2090 \\
& & METEOR $\uparrow$ & 0.2229 & 0.2248 & 0.2244 & 0.1980 & \textbf{0.2248} & 0.2244 & 0.1980 \\
\cmidrule{2-10}
& \multirow{2}{*}{Temporal} & ROUGE-1 $\uparrow$ & 0.3471 & \textbf{0.3497} & 0.3483 & 0.2592 & 0.3497 & 0.3483 & 0.2592 \\
&  & ROUGE-L $\uparrow$ & 0.2312 & 0.2317 & 0.2299 & 0.1671 & \textbf{0.2317} & 0.2299 & 0.1671 \\
& & METEOR $\uparrow$ & 0.2102 & 0.2119 & 0.2111 & 0.1603 & \textbf{0.2119} & 0.2111 & 0.1603 \\
\midrule
\multirow{6}{*}{\shortstack[l]{{\bf Personalized}\\{\bf Topic Writing}}} & 
\multirow{2}{*}{User} & ROUGE-1 $\uparrow$ & 0.1685 & 0.1745 & 0.1766  & 0.1665 & 0.1742 & \textbf{0.1767} & 0.1653 \\
& & ROUGE-L $\uparrow$ & 0.1196 & 0.1201 & \textbf{0.1226}  & 0.1129 & 0.1200 & 0.1225 &  0.1129 \\
& & METEOR $\uparrow$ & 0.0961 & 0.0998 & \textbf{0.1010} & 0.0960 & 0.0999 & 0.1009 & 0.0960 \\
\cmidrule{2-10}
& \multirow{2}{*}{Temporal} & ROUGE-1 $\uparrow$ & 0.1682 & 0.1784 & \textbf{0.1835} & 0.1701 & 0.1783 & 0.1833 & 0.1702 \\
&  & ROUGE-L $\uparrow$ & 0.1192 & 0.1238 & \textbf{0.1264} & 0.1148 & 0.1239 & 0.1262 & 0.1147  \\
& & METEOR $\uparrow$ & 0.0970 & 0.1024 & \textbf{0.1059} & 0.0994 & 0.1024 & 0.1059 & 0.0994 \\
\midrule
\multirow{6}{*}{\shortstack[l]{{\bf Personalized}\\{\bf Review Writing}}} & 
\multirow{2}{*}{User} & ROUGE-1 $\uparrow$ & 0.2063 & \textbf{0.2199} & 0.2046 & 0.1984 & 0.2197 & 0.2046 & 0.1985 \\
&  & ROUGE-L $\uparrow$ & 0.1350 & \textbf{0.1414} &  0.1312 & 0.1225 & 0.1413 & 0.1312 & 0.1225 \\
& & METEOR $\uparrow$ & 0.1055 & \textbf{0.1141} & 0.1059 & 0.1032 & 0.1141 & 0.1059 & 0.1032 \\
\cmidrule{2-10}
& \multirow{2}{*}{Temporal} & ROUGE-1 $\uparrow$ & 0.2074 & \textbf{0.226}4 & 0.2171 & 0.2132 & 0.2253 & 0.2171 & 0.2133  \\
&  & ROUGE-L $\uparrow$ & 0.1362 & \textbf{0.1448} & 0.1340 & 0.1225 & 0.1448 & 0.1340 &  0.1224 \\
& & METEOR $\uparrow$ & 0.1076 & \textbf{0.1188} & 0.1151 & 0.1138 & 0.1185 & 0.1151 & 0.1138 \\
\midrule
\multirow{6}{*}{\shortstack[l]{{\bf Personalized}\\{\bf Email Writing}}} &
\multirow{2}{*}{User} & ROUGE-1 $\uparrow$ & 0.1976 & 0.3239 & \textbf{0.3331} & 0.3055 & 0.3243 & 0.3331 & 0.3047 \\
& & ROUGE-L $\uparrow$ & 0.1530 & 0.2866 & 0.2937 & 0.2713 & 0.2867 & \textbf{0.2938} &  0.2713 \\
& & METEOR $\uparrow$ & 0.1477 & 0.2684 & 0.2762 & 0.2506 & 0.2684 & \textbf{0.2762} & 0.2506 \\
% \addlinespace
\cmidrule{2-10}
& \multirow{2}{*}{Temporal} & ROUGE-1 $\uparrow$ & 0.2389 & 0.4008 & 0.4016 & 0.3827 & 0.4020 & \textbf{0.4020} & 0.3835 \\
&  & ROUGE-L $\uparrow$ & 0.2000 & \textbf{0.3647} & 0.3638 & 0.3425 & 0.3642 & 0.3635 & 0.3419 \\
& & METEOR $\uparrow$ & 0.1930 & 0.3381 & \textbf{0.3403} & 0.3266 & 0.3381 & 0.3403 & 0.3266 \\
\bottomrule
\end{tabular}
\end{adjustbox}
\caption{Personalized long-text generation results on a fine-tuned language model: FlanT5-base on the validation set. Note k denotes the retrieved item count for a specific user for personalizing the generation.
}
\label{table:finetuning-validation-results}
\end{table*}

\begin{table*}[h!]
\centering
\scalebox{0.8}{
\begin{tabular}{ll ccc c ccc}
\toprule
& & 
\multicolumn{3}{c}{{\bf LLaMA2}} & 
\multicolumn{3}{c}{{\bf GPT-3.5}}
\\
\cmidrule(r){3-5}
\cmidrule(r){6-8}
{\bf Benchmark} & {\bf Metric} & {k=1} & {k=2} &{k=4} & {k=1} & {k=2} &{k=4}
\\
\midrule
\multirow{2}{*}{\shortstack[l]{{\bf Personalized}\\{\bf Abstract Generation}}} & 
ROUGE-1 $\uparrow$ & 0.3175 & 0.3582 & \textbf{0.3651} & 0.3805 & 0.3801 & \textbf{0.3827}  \\
& ROUGE-L $\uparrow$ & 0.1720 & 0.1921 & \textbf{0.1970} & 0.2132 & 0.2134 & \textbf{0.2152}  \\
\midrule
\multirow{2}{*}{\shortstack[l]{{\bf Personalized}\\{\bf Review Writing}}} &
ROUGE-1 $\uparrow$ & \textbf{0.2885} & 0.2875 & 0.2798 & 0.2905 & \textbf{0.2939} & 0.2905\\
& ROUGE-L $\uparrow$ & \textbf{0.1384} & 0.1373 & 0.1361 & 0.1446 & \textbf{0.1469} & 0.1464\\
\midrule
\multirow{2}{*}{\shortstack[l]{{\bf Personalized}\\{\bf Email Writing}}} &
ROUGE-1 $\uparrow$ & 0.2852 & 0.2955 & \textbf{0.3016} & - & - & - \\
& ROUGE-L $\uparrow$ & 0.2257 & 0.2402  &  \textbf{0.2512} & - & - & - \\
\bottomrule
\end{tabular}
}
\caption{Temporal-based setting with recency using LLaMA2-7B and GPT-3.5 models on the validation set}
\label{tab:recency-llama}
\end{table*}

\begin{table*}[!t]
    \centering
    \scalebox{0.80}{
        \begin{tabular}{ll cccc cc HH HHHHHH}
        \toprule
        & & \multicolumn{2}{c}{{Non-Personalized}} & 
        \multicolumn{2}{c}{{ Personalized-Recency}} &
        \multicolumn{2}{c}{{ k}} &
        \\
        \cmidrule(r){3-4}
        \cmidrule(lr){5-6}
        \cmidrule(lr){7-8}
        {{\bf Benchmark Dataset}}  & {{\bf Metric}} & \multicolumn{1}{c}{LLaMA2} & {GPT-3.5\;\;} & {\;\;LLaMA2\;} & {GPT-3.5} & 
        \;\;LLaMA2\;\; &
        \;\;GPT-3.5\;\; &
        \\\midrule        
        \multirow{3}{*}{\shortstack[l]{{\bf Personalized}\\{\bf Abstract Generation}}} & ROUGE-1 $\uparrow$ & 0.3515 & 0.3687 & \textbf{0.3624} & \textbf{0.3801} & 4 & 4 & \textbf{0.000} & 0.000 \\
        & ROUGE-L $\uparrow$ & 0.1885 & 0.2094 & \textbf{0.1952} & \textbf{0.2145} & 4 & 4 & \textbf{0.000} & 0.000 \\
        & METEOR $\uparrow$ & \textbf{0.2813} & 0.2314 & 0.2662 & \textbf{0.2442} & 4 & 4 & \textbf{0.000} & 0.000 \\
        \midrule
        \multirow{3}{*}{\shortstack[l]{{\bf Personalized}\\{\bf Review Writing}}} & ROUGE-1 $\uparrow$ & 0.2839 & 0.2841 & \textbf{0.2846} & \textbf{0.2920} & 1 & 2 & \textbf{0.000} & 0.000 \\
        & ROUGE-L $\uparrow$ & \textbf{0.1395} & 0.1413 & 0.1372 & \textbf{0.1473} & 1 & 2 & \textbf{0.000} & 0.000 \\
        & METEOR $\uparrow$ & 0.1772 & 0.1584 & \textbf{0.1868} & \textbf{0.1669} & 1 & 2 & \textbf{0.000} & 0.000 \\
        \midrule
        \multirow{3}{*}{\shortstack[l]{{\bf Personalized}\\{\bf Email Writing}}} 
        & ROUGE-1 $\uparrow$ & 0.1825 & - & \textbf{0.3264} & - & 4 & - & \textbf{0.000} & - \\
        & ROUGE-L $\uparrow$ & 0.1159 & - & \textbf{0.2721} & - & 4 & - & \textbf{0.000} & - \\
        & METEOR $\uparrow$ & 0.1622 & - & \textbf{0.3126} & - & 4 & - & \textbf{0.000} & - \\   
        \bottomrule
        \end{tabular}
    }
    \caption{Temporal-based separation setting with recency using LLaMA2-7B and GPT-3.5 models on the test set}
\label{tab:recency-test-setting}
\end{table*}

\begin{table*}[h!]
\centering
\scalebox{0.8}{
\begin{tabular}{ll cccc c}
\toprule
& & 
\multicolumn{1}{c}{{\bf Non-Personalized}} & 
\multicolumn{1}{c}{{\bf Non-Personalized}} &
\multicolumn{1}{c}{{\bf Personalized}} &
\multicolumn{1}{c}{{\bf Personalized}} &
\\
{\bf Benchmark} & {\bf Metric} & {} &{\bf Random} & {\bf Untuned} 
& {\bf Retrieved} 
\\
\midrule
\multirow{2}{*}{\shortstack[l]{{\bf Personalized}\\{\bf Abstract Generation}}} & 
ROUGE-1 $\uparrow$ & 0.3872 & 0.3824 & 0.3881 & 0.4157  \\
& ROUGE-L $\uparrow$ & 0.2207 & 0.217 & 0.220 & 0.2416  \\
\midrule
\multirow{2}{*}{\shortstack[l]{{\bf Personalized}\\{\bf Topic Writing}}} &
ROUGE-1 $\uparrow$ & 0.2773 & 0.2557 & 0.269 & 0.2904 \\
& ROUGE-L $\uparrow$ & 0.1397 &0.1208 & 0.125 & 0.1368\\
\midrule
\multirow{2}{*}{\shortstack[l]{{\bf Personalized}\\{\bf Review Writing}}} &
ROUGE-1 $\uparrow$ & 0.2746 & 0.2731 & 0.282 & 0.2857\ \\
& ROUGE-L $\uparrow$ & 0.1384 & 0.1373 & 0.141  &  0.1450 \\
\bottomrule
\end{tabular}
}
\caption{User-based separation setting baselines using GPT-3.5 model on the test set (k=1)}
\label{tab:additional-baselines-user-setting}
\end{table*}

\begin{table*}[h!]
\centering
\scalebox{0.8}{
\begin{tabular}{ll cccc c}
\toprule
& & 
\multicolumn{1}{c}{{\bf Non-Personalized}} & 
\multicolumn{1}{c}{{\bf Non-Personalized}} &
\multicolumn{1}{c}{{\bf Personalized}} &
\multicolumn{1}{c}{{\bf Personalized}} &
\\
{\bf Benchmark} & {\bf Metric} & {} &{\bf Random} & {\bf Untuned} 
& {\bf Retrieved} 
\\
\midrule
\multirow{2}{*}{\shortstack[l]{{\bf Personalized}\\{\bf Abstract Generation}}} & 
ROUGE-1 $\uparrow$ & 0.3687 & 0.3821 & 0.3865 & 0.3830  \\
& ROUGE-L $\uparrow$ & 0.2094 & 0.217 & 0.215 & 0.2167  \\
\midrule
\multirow{2}{*}{\shortstack[l]{{\bf Personalized}\\{\bf Topic Writing}}} &
ROUGE-1 $\uparrow$ & 0.2841 & 0.260 & 0.271 & 0.2904 \\
& ROUGE-L $\uparrow$ & 0.1413 & 0.122 & 0.217 & 0.1363\\
\midrule
\multirow{2}{*}{\shortstack[l]{{\bf Personalized}\\{\bf Review Writing}}} &
ROUGE-1 $\uparrow$ & 0.2836 & 0.2601 & 0.289 & 0.2958 \ \\
& ROUGE-L $\uparrow$ & 0.1422 & 0.122 & 0.144  &  0.1479 \\
\bottomrule
\end{tabular}
}
\caption{Temporal-based separation setting baselines using GPT-3.5 model on the test set (k=1)}
\label{tab:additional-baselines-temporal-setting}
\end{table*}

\begin{table*}[h!]
\centering
\scalebox{0.8}{
\begin{tabular}{ll cccc c}
\toprule
& & 
\multicolumn{1}{c}{{\bf Non-Personalized}} & 
\multicolumn{1}{c}{{\bf Non-Personalized}} &
\multicolumn{1}{c}{{\bf Personalized}} &
\multicolumn{1}{c}{{\bf Personalized}} &
\\
{\bf Benchmark} & {\bf Metric} & {} &{\bf Random} & {\bf Untuned} 
& {\bf Retrieved} 
\\
\midrule
\multirow{2}{*}{\shortstack[l]{{\bf Personalized}\\{\bf Abstract Generation}}} & 
ROUGE-1 $\uparrow$ & 0.3508 & 0.3094 & 0.3193 & 0.3511  \\
& ROUGE-L $\uparrow$ & 0.1923 & 0.1707 & 0.1757 & 0.1961  \\
\midrule
\multirow{2}{*}{\shortstack[l]{{\bf Personalized}\\{\bf Topic Writing}}} &
ROUGE-1 $\uparrow$ & 0.2184 & 0.1806 & 0.1904 & 0.2540 \\
& ROUGE-L $\uparrow$ & 0.1109 &0.0959 & 0.1003 & 0.1258\\
\midrule
\multirow{2}{*}{\shortstack[l]{{\bf Personalized}\\{\bf Review Writing}}} &
ROUGE-1 $\uparrow$ & 0.2766 & 0.2771 & 0.2848 & 0.2870\ \\
& ROUGE-L $\uparrow$ & 0.1351 & 0.1335 & 0.1360  &  0.1384 \\

\midrule
\multirow{2}{*}{\shortstack[l]{{\bf Personalized}\\{\bf Email Writing}}} &
ROUGE-1 $\uparrow$ & 0.1773 & 0.1623 & 0.1773 & 0.3551\ \\
& ROUGE-L $\uparrow$ & 0.1111 & 0.1026 & 0.1111  &  0.3018 \\
\bottomrule
\end{tabular}
}
\caption{User-based separation setting baselines using LLaMA2-7B model on the test set (k=1)}
\label{tab:additional-baselines-user-setting-llama}
\end{table*}

\begin{table*}[h!]
\centering
\scalebox{0.8}{
\begin{tabular}{ll cccc c}
\toprule
& & 
\multicolumn{1}{c}{{\bf Non-Personalized}} & 
\multicolumn{1}{c}{{\bf Non-Personalized}} &
\multicolumn{1}{c}{{\bf Personalized}} &
\multicolumn{1}{c}{{\bf Personalized}} &
\\
{\bf Benchmark} & {\bf Metric} & {} &{\bf Random} & {\bf Untuned} 
& {\bf Retrieved} 
\\
\midrule
\multirow{2}{*}{\shortstack[l]{{\bf Personalized}\\{\bf Abstract Generation}}} & 

ROUGE-1 $\uparrow$ & 0.3515 & 0.3096 & 0.3129 & 0.3285  \\
& ROUGE-L $\uparrow$ & 0.1885 & 0.1670 & 0.1693 & 0.1780  \\

\midrule
\multirow{2}{*}{\shortstack[l]{{\bf Personalized}\\{\bf Topic Writing}}} &

ROUGE-1 $\uparrow$ & 0.2193 & 0.1861 & 0.1922 & 0.2487 \\
& ROUGE-L $\uparrow$ & 0.1108 & 0.0975 & 0.1002 & 0.1226\\

\midrule
\multirow{2}{*}{\shortstack[l]{{\bf Personalized}\\{\bf Review Writing}}} &

ROUGE-1 $\uparrow$ & 0.2839 & 0.2789 & 0.2879 & 0.2877 \\
& ROUGE-L $\uparrow$ & 0.1395 & 0.1348 & 0.1382  &  0.1384 \\

\midrule
\multirow{2}{*}{\shortstack[l]{{\bf Personalized}\\{\bf Email Writing}}} &

ROUGE-1 $\uparrow$ & 0.1825 & 0.1725 & 0.2723 & 0.2986\ \\
& ROUGE-L $\uparrow$ & 0.1159 & 0.1093 & 0.2141  &  0.2373 \\

\bottomrule
\end{tabular}
}
\caption{Temporal-based separation setting baselines using LLaMA2-7B model on the test set (k=1)}
\label{tab:additional-baselines-temporal-setting-llama}
\end{table*}

\section{Performance on Validation \& Gains}
\label{appendix: val-performance}
This section reports the results of experiments on the validation set. Table \ref{table:zero-shot-validation-results-gpt} shows the results of the zero-shot evaluation of GPT-3.5 on the user and temporal settings on the validation set. The Personalized Abstract Generation benchmark exhibits the highest increase just as it did in the test set results displayed in Table \ref{tab:zero-shot-test-results-user-setting}-\ref{tab:zero-shot-test-results-time-setting}. Table \ref{table:zero-shot-validation-results-LLaMA7} shows the results of the zero-shot evaluation of Llama-7B on the user and temporal settings on the validation set. 
% The Personalized Email Completion benchmark demonstrates the highest increase which again is the same case as in the test set results given in Table \ref{tab:zero-shot-test-results-user-setting}-\ref{tab:zero-shot-test-results-time-setting}. 

Table \ref{table:finetuning-validation-results} presents the validation results for fine-tuning with FlanT5-base in both user and temporal settings. 
Tables \ref{table:gain-results-fine-tuning}, \ref{tab:gain-results-user-setting} and \ref{tab:gain-results-temporal-setting} capture the percentage gain over the non-personalized results.

\begin{table*}[!t]
\centering
\scalebox{0.8}{
\begin{tabular}{llccc c}
\toprule
& & 
\multicolumn{1}{c}{{\bf Personalized}} & 
\multicolumn{1}{c}{{\bf Personalized}} &
\multicolumn{1}{c}{{\bf Personalized}} &
\\
{\bf Benchmark} & {\bf Metric} & {\bf LLaMA2 Gain (\%)} & {\bf GPT-3.5 Gain (\%)} 
& {\bf Overall Gain (\%)} 
\\
\midrule
\multirow{3}{*}{\shortstack[l]{{\bf Personalized}\\{\bf Abstract Generation}}} & 

ROUGE-1 $\uparrow$ & 11.67\% & 7.36\% & 9.67\%  \\
& ROUGE-L $\uparrow$ & 14.72\% & 9.47\% & 12.09\%  \\
& METEOR $\uparrow$ & 2.06\% & 12.85\% & 7.45\%  \\
\midrule
\multirow{3}{*}{\shortstack[l]{{\bf Personalized}\\{\bf Topic Writing}}} &

ROUGE-1 $\uparrow$ & 16.30\% & 4.72\% & 10.51\% \\
& ROUGE-L $\uparrow$ & 13.44\% & -2.08\% & 5.68\% \\
& METEOR $\uparrow$ & 22.57\% & 35.27\% & 28.92\% \\
\midrule
\multirow{3}{*}{\shortstack[l]{{\bf Personalized}\\{\bf Review Writing}}} &

ROUGE-1 $\uparrow$ & 3.62\% & 4.04\% & 3.83\% \\
& ROUGE-L $\uparrow$ & 1.55\% & 4.77\%  & 3.16\%  \\
& METEOR $\uparrow$ & 15.12\% & 5.12\% & 10.12\%  \\
\midrule
\multirow{3}{*}{\shortstack[l]{{\bf Personalized}\\{\bf Email Writing}}} &

ROUGE-1 $\uparrow$ & 96.84\% & - & 96.84\%\\
& ROUGE-L $\uparrow$ & 169.40\% & - & 169.40\%\\
& METEOR $\uparrow$ & 117.76\% & - & 117.76\%\\
\bottomrule
\end{tabular}
}
\caption{Percent gain between personalized and non-personalized results for the user-based separation setting on LLaMA2 and GPT-3.5 models in the zero-shot setting . 
}
\label{tab:gain-results-user-setting}
\end{table*}

\begin{table*}[!t]
    \centering
    \scalebox{0.80}{
\begin{tabular}{llccc c}
\toprule
& & 
\multicolumn{1}{c}{{\bf Personalized}} & 
\multicolumn{1}{c}{{\bf Personalized}} &
\multicolumn{1}{c}{{\bf Personalized}} &
\\
{\bf Benchmark} & {\bf Metric} & {\bf LLaMA2 Gain (\%)} & {\bf GPT-3.5 Gain (\%)} 
& {\bf Overall Gain (\%)} 
\\
        \midrule
        \multirow{3}{*}{\shortstack[l]{{\bf Personalized}\\{\bf Abstract Generation}}} & 
        ROUGE-1 $\uparrow$ & 5.15\% & 3.88\%  & 4.52\% \\
                                         & ROUGE-L $\uparrow$ & 6.33\% & 3.49\%  & 4.91\% \\
                                         & METEOR $\uparrow$  & -3.47\% & 7.05\%  & 1.78\% \\
        \midrule
        \multirow{3}{*}{\shortstack[l]{{\bf Personalized}\\{\bf Topic Writing}}} &
                                            ROUGE-1 $\uparrow$ & 13.41\% & 2.22\%  & 7.81\% \\
                                         & ROUGE-L $\uparrow$ & 10.65\% & -3.53\%  & 3.56\% \\
                                         & METEOR $\uparrow$  & 20.09\% & 31.12\%  & 25.61\% \\
        \midrule
        \multirow{3}{*}{\shortstack[l]{{\bf Personalized}\\{\bf Review Writing}}} &
        ROUGE-1 $\uparrow$ & 0.24\% & 4.30\%  & 2.274\% \\
                                         & ROUGE-L $\uparrow$ & -2.58\% & 4.01\%  & 0.713\% \\
                                         & METEOR $\uparrow$  & 9.82\% & 5.76\%  & 7.79\% \\
        \midrule
        \multirow{3}{*}{\shortstack[l]{{\bf Personalized}\\{\bf Email Writing}}} &
        ROUGE-1 $\uparrow$ & 71.34\% & -  & 71.34\% \\
                                         & ROUGE-L $\uparrow$ & 121.13\% & -  & 121.13\% \\
                                         & METEOR $\uparrow$  & 84.77\% & -  & 84.77\% \\
        \bottomrule
        \end{tabular}
    }
    \caption{Percent gain between personalized and non-personalized results for the temporal setting on LLaMA2 and GPT-3.5 models in the zero-shot setting.
    }
    \label{tab:gain-results-temporal-setting}
\end{table*}

\begin{table*}[!t]
    \centering
    \scalebox{0.80}{
\begin{tabular}{llccc c}
\toprule
& & 
\multicolumn{1}{c}{{\bf Personalized}} & 
\multicolumn{1}{c}{{\bf Personalized}} &
\\
{\bf Benchmark} & {\bf Metric} & {\bf User Setting(\%)} & {\bf Temporal Setting (\%)} 
\\
        \\
        \midrule
        \multirow{3}{*}{\shortstack[l]{{\bf Personalized}\\{\bf Abstract Generation}}} & 
        
                                        ROUGE-1 $\uparrow$ & 0.44\% & 0.75\%  \\
                                         & ROUGE-L $\uparrow$ & -0.08\% & 4.02\%  \\
                                         & METEOR $\uparrow$  & 0.67\% & 1.83\%  \\
        \midrule
       
        \multirow{3}{*}{\shortstack[l]{{\bf Personalized}\\{\bf Topic Writing}}} &
                                            ROUGE-1 $\uparrow$ & 6.06\% & 10.43\% \\
                                         & ROUGE-L $\uparrow$ & 4.60\% & 6.74\%  \\
                                         & METEOR $\uparrow$  & 7.34\% & 10.18\%  \\
        \midrule
       
        \multirow{3}{*}{\shortstack[l]{{\bf Personalized}\\{\bf Review Writing}}} &
        ROUGE-1 $\uparrow$ & 5.60\% & 8.78\% \\
                                         & ROUGE-L $\uparrow$ & 3.21\% & 5.24\%  \\
                                         & METEOR $\uparrow$  & 6.89\% & 9.89\%  \\
        \midrule
        
        \multirow{3}{*}{\shortstack[l]{{\bf Personalized}\\{\bf Email Writing}}} &
        ROUGE-1 $\uparrow$ & 78.4\% & 69.7\% \\
                                         & ROUGE-L $\uparrow$ & 101.8\% & 86.2\%   \\
                                         & METEOR $\uparrow$  & 82.4\% & 77.0\%  \\
        \bottomrule
        \end{tabular}
    }
    \caption{Percent gain between personalized and non-personalized results for the user and temporal separation settings in fine-tuning using FlanT5-base.
    }
    \label{table:gain-results-fine-tuning}
\end{table*}

\section{Additional Experiments}
\subsection{Additional Baselines and Retriever}
\label{appendix:additional-baselines}
We experiment with additional two additional baselines and an additional retriever to establish comparative performance. The Non-Personalized Random baseline randomly retrieves a profile from the amalgamated collection of all users' profiles. The Personalized Untuned baseline randomly retrieves from the target user's profile set. The results of these experiments are presented in Tables \ref{tab:additional-baselines-user-setting-llama} and Table \ref{tab:additional-baselines-user-setting} for user setting and Table \ref{tab:additional-baselines-temporal-setting-llama} Table \ref{tab:additional-baselines-temporal-setting} for the temporal setting.
We also use Recency as an additional retriever. The Recency retriever operates by selecting the most temporally recent profiles for each user in the dataset. Intuitively, a user's more recent profiles are likely to better capture their latest interests, language patterns, and personal context compared to older profiles. As such, retrieving these latest profiles can serve as a strong personalization signal by prioritizing the most up-to-date user representations available. The recency heuristic provides a simple yet effective baseline for personalization in the temporal analysis setting.

% \subsection{Additional Experiments on the framework}
\subsection{Summarization Experiments}
Upon establishing our baselines we perform an additional set of experiments, on the $\phi_{p}$ function to incorporate more information into the prompt for better personalization and generation. 

\begin{itemize}
    \item \textbf{Summarization Only}
    This approach generates a summary of each document retrieved from $\mathcal{R}$. The summaries along with the original prompt $x_{i}$ is given to the $\phi_{p}$, which follows the similar procedure of the summaries and the input prompt to $\bar{x_i}$. More formally:
    \begin{align}
    s_{j} &= \textsc{llm}(r_j), \quad \forall r_j \in \mathcal{R}(\phi_q(x_i), P_u, k)\\
    S_u &= \{\ldots,s_{j},\ldots\}
    \end{align}
where $S_u$ is the set of summarized text for user $u$ where $s_j \in S_u$ refers to the summary of $r_j \in \mathcal{R}(\phi_q(x_i), P_u, k)$.

    \begin{align}
    \bar{x_i} &= \phi_p(x_i,S_u) \\
    &= \phi_p(x_i, \{s_{i1},\ldots,s_{ik}\})\\
    \bar{y_i} &= \textsc{llm}(x_i)
    \end{align}

    here $\bar{y_i}$ is the final output generated by the LLM.

   The results of the zero-shot experiments on our benchmark dataset for the summarization task are presented in Table \ref{table:zero-shot-validation-results-gpt-summarization-only}. We evaluated the performance using two different retrievers, namely BM25 and Contriever, with each retriever configured to retrieve the top $k$ relevant documents, where $k$ was set to 1, 2, and 4. The experiments were conducted on the validation sets for all tasks in our benchmark. Across all datasets, our results demonstrate a significant improvement over the baseline, where no personalized documents from the user's profile were retrieved. Notably, when comparing to the results presented in Table \ref{table:zero-shot-validation-results-gpt} which contain the validation set performance using the framework described in Section \ref{sec:3}, we observe significant improvements on two specific tasks: review generation and topic generation. These gains underscore the efficacy of our proposed approach, particularly for these two challenging datasets.

   \begin{table*}[h!]
\centering
\begin{adjustbox}{max width=\textwidth}
\begin{tabular}{lll c ccc ccc}
\toprule

\multirow{2}{*}{\bf Personalized Long-Text} &  \multirow{2}{*}{\bf Setting} & \multirow{2}{*}{\bf Metric} & & \multicolumn{3}{c}{BM25} & \multicolumn{3}{c}{Contriever} \\
\cmidrule(r){5-7} \cmidrule(l){8-10}
{\bf Benchmark Data} & & & No Personal. & $k=1$ & $k=2$ & $k=4$ & $k=1$ & $k=2$ & $k=4$ \\
\midrule
\multirow{4}{*}{\shortstack[l]{{\bf Personalized}\\{\bf Abstract Generation}}} &
\multirow{2}{*}{User} & ROUGE-1 $\uparrow$ & 0.3727 & 0.3912 & 0.3957 & 0.3980 & 0.3903 & 0.3962 & 0.3973 \\
& & ROUGE-L $\uparrow$ & 0.2102 & 0.2196 & 0.2203 & 0.2193 & 0.2191 & 0.2205 & 0.2198 \\

\cmidrule{2-10}
& \multirow{2}{*}{Temporal} & ROUGE-1 $\uparrow$ & 0.3727 & 0.3737 & 0.3753 & 0.3781 & 0.3728 & 0.3755 & 0.3782 \\
&  & ROUGE-L $\uparrow$ & 0.2101 & 0.2158 & 0.2171 & 0.2171 & 0.2157 & 0.2165 & 0.2168 \\

\midrule

\multirow{4}{*}{\shortstack[l]{{\bf Personalized}\\{\bf Topic Writing}}} & 
\multirow{2}{*}{User} & ROUGE-1 $\uparrow$ & 0.25813 & \textbf{0.2936} & 0.2891 & 0.2844 & 0.2931 & 0.29170 &  0.2860 \\
& & ROUGE-L $\uparrow$ & 0.1251 & \textbf{0.1362} & 0.1353 & 0.1342 & 0.1360 & 0.1355 & 0.1346 \\

\cmidrule{2-10}
& \multirow{2}{*}{Temporal} & ROUGE-1 $\uparrow$ & 0.26051 & \textbf{0.294}3 & 0.2876 & 0.2854 & 0.2942 & 0.2906 & 0.2871 \\
&  & ROUGE-L $\uparrow$ & 0.1275 & \textbf{0.1361} & 0.1350 & 0.1355 & 0.1367 & 0.1362 & 0.1351 \\

\midrule
\multirow{4}{*}{\shortstack[l]{{\bf Personalized}\\{\bf Review Writing}}} & 
\multirow{2}{*}{User} & ROUGE-1 $\uparrow$ & 0.2768 & 0.3046 & 0.3046 & 0.3012 & \textbf{0.3047} & 0.3042 & 0.2999 \\
&  & ROUGE-L $\uparrow$ & 0.1397 & 0.1440 & 0.1447 & 0.1443 & 0.1442 & \textbf{0.1449} & 0.1438 \\

\cmidrule{2-10}
& \multirow{2}{*}{Temporal} & ROUGE-1 $\uparrow$ & 0.2829 & 0.3066 & 0.3067 & 0.3026 & \textbf{0.3073} & 0.3050 & 0.000 \\
&  & ROUGE-L $\uparrow$ & 0.1412 & 0.1446 & 0.1448 & 0.1439 & \textbf{0.1450} & 0.1450 & 0.000 \\

\bottomrule
\end{tabular}
\end{adjustbox}
\caption{The zero-shot personalized results using GPT-3.5 model for user- and temporal settings for \textsc{Summarization Only}.Best results are bold}

\label{table:zero-shot-validation-results-gpt-summarization-only}
\end{table*}

    \item \textbf{Stylistic Extraction Only}:
    This approach first seeks to extract the stylistic elements of each user, and then leverage it in the prompting function along with the retrieved profiles for the final generation. More formally,
    \begin{align}
R_{u} &= \{\ldots,r_{j},\ldots\}, \quad \forall r_j \in \mathcal{R}(\phi_q(x_i), P_u, k)\\
 {l_{u}} &= \textsc{llm}(R_{u})
\end{align}
 where $R_u$ is the set of retrieved text for user $u$. $l_u$ is the linguistic properties generated for each user \textit{u}.
  \begin{align}
\bar{x_i} &= \phi_p(x_i,\mathcal{R}(\phi_q(x_i), P_u, k)),l_u) \\
&= \phi_p(x_i, \{r_{i1},\ldots,r_{ik}\},l_u)\\
\bar{y_i} &= \textsc{llm}(x_i)
\end{align}
here $\bar{y_i}$ is the final output generated by the LLM.
\end{itemize}

The results of the zero-shot experiments on our benchmark dataset for the style task are presented in Table \ref{table:zero-shot-validation-results-gpt-style-only}. We evaluated the performance using two different retrievers, namely BM25 and Contriever, with each retriever configured to retrieve the top $k$ relevant documents, where $k$ was set to 1, 2, and 4. The experiments were conducted on the validation sets for all tasks in our benchmark. Across all datasets, our results demonstrate a significant improvement over the baseline, where no personalized documents from the user's profile were retrieved.

\begin{table*}[ht]
\centering
\begin{adjustbox}{max width=\textwidth}
\begin{tabular}{lll c ccc ccc}
\toprule

\multirow{2}{*}{\bf Personalized Long-Text} &  \multirow{2}{*}{\bf Setting} & \multirow{2}{*}{\bf Metric} & & \multicolumn{3}{c}{BM25} & \multicolumn{3}{c}{Contriever} \\
\cmidrule(r){5-7} \cmidrule(l){8-10}
{\bf Benchmark Data} & & & No Personal. & $k=1$ & $k=2$ & $k=4$ & $k=1$ & $k=2$ & $k=4$ \\
\midrule
\multirow{4}{*}{\shortstack[l]{{\bf Personalized}\\{\bf Abstract Generation}}} &
\multirow{2}{*}{User} & ROUGE-1 $\uparrow$ & 0.3727 & 0.3928 & \textbf{0.3928} & 0.3925 & 0.3925 & 0.3928 & 0.3924 \\
& & ROUGE-L $\uparrow$ & 0.32121 & 0.2176 & \textbf{0.2178} & 0.2175 & 0.2178 & 0.2180 & 0.2175 \\

\cmidrule{2-10}
& \multirow{2}{*}{Temporal} & ROUGE-1 $\uparrow$ & 0.3727  & 0.3828 & 0.3834 & 0.3833 & 0.3823 & \textbf{0.3838} & 0.3835  \\
&  & ROUGE-L $\uparrow$ & 0.2101 &  0.2109 & 0.2108 & 0.2105 & 0.2102 & \textbf{0.2112} & 0.21087 \\

\midrule

\multirow{4}{*}{\shortstack[l]{{\bf Personalized}\\{\bf Topic Writing}}} & 
\multirow{2}{*}{User} & ROUGE-1 $\uparrow$ & 0.2581 & 0.2812 & 0.2827 & \textbf{0.2838} & 0.2825 & 0.2817 & 0.2816 \\
& & ROUGE-L $\uparrow$ & 0.1251 & 0.1303 & 0.1307 & \textbf{0.1307} & 0.1310 & 0.1307 & 0.1306 \\

\cmidrule{2-10}
& \multirow{2}{*}{Temporal} & ROUGE-1 $\uparrow$ & 0.2605 & 0.2805 & \textbf{0.2819} & 0.2811 & 0.2810 & 0.2796 & 0.2812 \\
&  & ROUGE-L $\uparrow$ & 0.1275 & 0.1309 & \textbf{0.1312} & 0.1311 & 0.1312 & 0.1302 & 0.1310 \\

\midrule
\multirow{4}{*}{\shortstack[l]{{\bf Personalized}\\{\bf Review Writing}}} & 
\multirow{2}{*}{User} & ROUGE-1 $\uparrow$ & 0.2768 & 0.3073 & 0.3075 & 0.3081 & 0.3072 & 0.3068 & \textbf{0.3090} \\
&  & ROUGE-L $\uparrow$ & 0.1397 & 0.1463 & 0.1468 & 0.1468 & 0.1462 & 0.1464 & \textbf{0.1468} \\

\cmidrule{2-10}
& \multirow{2}{*}{Temporal} & ROUGE-1 $\uparrow$ & 0.2829 & 0.3134 & 0.3126 & \textbf{0.3138} & 0.3119 & 0.3108 & 0.3111 \\
&  & ROUGE-L $\uparrow$ & 0.1412 & 0.1475 & 0.1475 & \textbf{0.1477} & 0.1474 &  0.1469 & 0.1467 \\

\bottomrule
\end{tabular}
\end{adjustbox}
\caption{The zero-shot personalized results using GPT-3.5 model for user and temporal settings for \textsc{Style Only}.The best results are highlighted in bold.}
\label{table:zero-shot-validation-results-gpt-style-only}
\end{table*}
\end{document}